# A Survey on Generative Adversarial Networks: Variants, Applications, and Training

Abdul Jabbar, Xi Li[*], and Bourahla Omar
College of Computer Science, Zhejiang University, Hangzhou, China;
Emails: {jabbar, xilizju, obourahla}@zju.edu.cn

**ABSTRACT**

The Generative Models have gained considerable attention in the field of unsupervised learning via a new and practical framework called Generative Adversarial Networks (GAN) due to its outstanding data generation capability. Many models of GAN have proposed, and several practical applications emerged in various domains of computer vision and machine learning. Despite GAN's excellent success, there are still obstacles to stable training. The problems are due to Nash-equilibrium, internal covariate shift, mode collapse, vanishing gradient, and lack of proper evaluation metrics. Therefore, stable training is a crucial issue in different applications for the success of GAN. Herein, we survey several training solutions proposed by different researchers to stabilize GAN training. We survey, (I) the original GAN model and its modified classical versions, (II) detail analysis of various GAN applications in different domains, (III) detail study about the various GAN training obstacles as well as training solutions. Finally, we discuss several new issues as well as research outlines to the topic.

## 1  Introduction

Most of the techniques used in Artificial Intelligence (AI) are supervised machine learning, whereas unsupervised learning is still a relatively unsolved research area. Recently, generative modeling, considerable with deep learning techniques, opened a new hope in the area of unsupervised learning, and Generative Adversarial Networks (GAN) is one of them. GAN is an example of generative models presented by (Goodfellow et al. 2014) [1]. GAN is the most common learning model in both semi-supervised and unsupervised learning. Theoretically, GAN takes a supervised learning approach to do unsupervised learning by generating fake or synthetic looking data. The essence of GAN can be summated as training of two networks simultaneously called the generator network denoted by *G* and the discriminator network indicated by *D*. *D* is a binary classifier learns to classify the real and generated data as genuinely as possible. In contrast, *G* confuses *D* by generating realistic data. These two networks section themselves, and eventually, *G* produces realistic data, and *D* gets better to predict the fake ones.

Practically, GAN have introduced many applications such as hand-written font generation [35 - 38], anime characters generation [45, 46], image blending [47, 50], image in-painting [51, 54], face aging [56 - 60], text synthesis [61 – 64, 126], human pose synthesis [65, 66], stenographic applications [67 - 69], image manipulation applications [70 - 73, 234], visual saliency prediction [74 - 76], object detection [77 - 80], 3D image synthesis [81, 82], medical applications [86, 87], facial makeup transfer [94 - 96], facial landmark d[1]etection [97, 98], image super-resolution [99 - 102], texture synthesis [103, 105, 106], sketch synthesis [107, 110, 111], image-to-image translation [34, 112 - 120], face frontal view generation [121-123, 125], language and speech synthesis [150, 151], music generation [152, 154, 155], video applications [168 - 171] in computer vision and graphics communities.

Although GANs have achieved some incredible results by producing stunningly realistic samples, however, it is still a difficult task to train the GAN by means of better stability. Stable GAN training is a crucial issue because both *G* and *D* need to optimize through alternating gradient descent (Alt-GD) or simultaneous gradient descent (Sim-GD). The architecture of GAN suffers due to several shortcomings, such as the Nash-equilibrium [191], internal covariate shift [192], mode collapse [193, 194], vanishing gradient [193], and lack of proper evaluation metrics [195]. Thus, several solutions such as feature matching [196], unrolled GAN [197], mini-batch discrimination [196], historical averaging [196], two time-scale update rule [198], hybrid model [72], self-attention GAN [199], relativistic GAN [203], label-smoothing [204], sampling GAN [205], proper optimizer [206 - 209], normalization techniques [193, 196, 211 - 216], add noise to inputs [217, 218], using labels [10, 56, 63, 109], alternative loss functions [220], gradient penalty [24], and cycle-consistency loss [113] have been introduced to stabilize GAN training. In this attempt, we explain the working of the GAN and its modified classical versions in detail. Also, we discuss how GANs have been applied to various practical applications, but not limited to the basic idea behind this survey, which is the training obstacle of the GAN and its potential solutions.

---

[*] Corresponding author, xilizju@zju.edu.cn

## 1.1 Structure of this survey

The structure of this survey paper is as follows: a concise introduction of the GAN and classical GAN-variants in Section 2. After that, an extensive comparative analysis of GAN variants, as Table 1 shows the summary of GAN variants reviewed in this section. Section 3 gives numerous expansions of the GAN practiced in different areas in artificial intelligence, as demonstrated in Table 3. In Section 4, we briefly survey several issues relative to the GAN stable training. At some relevant solutions that can improve the stability of the GAN during training. Finally, Section 5 concludes the survey paper. And, in particular, we address several new issues and potential future research directions on the topic.

Table 1: Summary of GAN variants reviewed in Section 2

| #   | Subject                                                                       | Pub.  | Year |
|-----|-------------------------------------------------------------------------------|-------|------|
| 1.  | Conditional Generative Adversarial Networks (CGAN) [15]                       | arXiv | 2014 |
| 2.  | Deep Convolution Generative Adversarial Networks (DCGAN) [17]                 | CoRR  | 2015 |
| 3.  | Laplacian Generative Adversarial Networks (LapGAN) [18]                       | NIPS  | 2015 |
| 4.  | Information Maximizing Generative Adversarial Networks (InfoGAN) [20]         | NIPS  | 2016 |
| 5.  | Energy-Based Generative Adversarial Networks (EBGAN) [22]                     | ICLR  | 2016 |
| 6.  | Wasserstein Generative Adversarial Networks (WGAN) [23]                       | ICML  | 2017 |
| 7.  | Boundary Equilibrium Generative Adversarial Networks (BEGAN) [25]             | CSLG  | 2017 |
| 8.  | Progressive-Growing Generative Adversarial Networks (PGGAN) [28]              | ICLR  | 2018 |
| 9.  | Big Generative Adversarial Networks (BigGAN) [29]                             | ICLR  | 2019 |
| 10. | Style-Based Generator Architecture for Generative Adversarial Networks [31]   | IEEE  | 2019 |

## 1.2 Relevance to other surveys and significance

In literature, several other relevant surveys (e.g., Gui et al. 2020 [2]; Wang et al. 2019 [3]; PAN et al. 2019 [4]; Hong et al. 2019 [5]; Hitawala. 2018 [6]; Creswell. 2018 [7]; Vuppuluri et al. 2017 [8]) of GAN have studied to investigate the recent trends, and their potential applications in different domains as Table 2 shows the list of most related survey papers. Among these surveys, the topic of the survey of Hong et al. 2019 [5], is most relevant to our survey. However, we have focused more on the classical models contributing to the significant "waves" in GAN, compared to the survey of Hong et al. 2019 [5], which rather concerns more on general GAN modeling. In addition, a comprehensive review of instability issues in GAN training and discussion on different solutions in detail to solve the training issues is focused. Practitioners, developers, and academics can use exhaustive survey literature of GAN, featuring; roots, key concepts, core techniques, and main modeling trends view-points. Specifically, a comparative analysis of GAN-variants to have a better understanding of their working. Second, we provide a detail review of several expansions of the GAN applied to different areas to have an in-depth knowledge of GAN achievements. Third, we offer a detail discussion on several problems associated with GAN training and their potential solutions. Fourth, we present a comparative analysis across GAN-training techniques to have a better understanding of applicability. Finally, we discuss several new issues for future research.

Table 2: Summary of related literature surveys

| #  | Subject                                                                           | Pub.  | Year |
|----|-----------------------------------------------------------------------------------|-------|------|
| 1. | A Review on Generative Adversarial Nets: Algorithms, Theory, and Applications [2] | arXiv | 2020 |
| 2. | Generative Adversarial Networks: A Survey and Taxonomy [3]                        | arXiv | 2019 |
| 3. | Recent Progress on Generative Adversarial Networks (GAN): A Survey [4]            | IEEE  | 2019 |
| 4. | How Generative Adversarial Nets and its Variants Work. An Overview of GAN [5]     | ACM   | 2019 |
| 5. | Comparative Study on Generative Adversarial Networks [6]                          | CoRR  | 2018 |
| 6. | Generative Adversarial Networks: An Overview [7]                                  | IEEE  | 2018 |
| 7. | Survey on Generative Adversarial Networks [8]                                     | IJERC | 2017 |

Table 3: Summary of GAN practical applications reviewed in Section 3

| Dom. | Subject with applied model names |
|---|---|
| Image | Handwritten font generation: DenseNet-CycleGAN [36], LS-CGAN [37], GlyphGAN [38] |
| | Anime characters generation: DRA-GAN [45], PS-GAN [46] |
| | Image blending: GP-GAN [47], GCC-GAN [50] |
| | Image in-painting: Ex-GAN [51], PG-GAN [54] |
| | Face aging: Age-CGAN [56], CAAE [57], IP-CGAN [58], Wavelet -GANs [59 - 60] |
| | Text synthesis: AttnGAN [61], StackGAN [62], ACGAN [63], TACGAN [64], SISGAN |
| | Human pose synthesis: $PG^2$[65], Deformable-GAN [66] |
| | Stenographic applications: S-GAN [67], SS-GAN [68], Stegano-GAN [69] |
| | Image manipulation: IGAN [70], TAGAN [71], IAN [72], AttGAN [73], DGAN [234] |
| | Visual saliency prediction: Sal-GAN [74], SalCapsule-CGAN [75], DSAL-GAN [76] |
| | Object detection: SeGAN [77], PGAN [78], MTGAN [79], GANDO [80] |
| | 3D image synthesis: 3DGAN [81], PrGAN [82] |
| | Medical applications: SeGAN [86], MedGAN [87] |
| | Facial makeup transfer: BGAN [94], PairedCycleGAN [95], DMTGAN [96] |
| | Face landmark detection: SAN [97], Expose-GAN [98] |
| | Image super-resolution: SR-GAN [99], ESR-GAN [100], SR-DGAN [101], T-GAN [102] |
| | Texture synthesis: MGAN [103], SGAN [105], PSGAN [106] |
| | Sketch synthesis: TGAN [107], Sketchy-GAN [110], CA-GAN [111] |
| | Image-to-image translation: Pix2pix [34], PAN [112], CycleGAN [113], Disco-GAN [114], Dual-GAN [115], StarGAN [116], UNIT [117], MUNIT [118], DRIT [119], DRIT++ [120] |
| | Face frontal view generation: DRGAN [121], TPGAN [122], FFGAN [123], FTGAN [125] |
| Audio | Speech and audio synthesis: Rank-GAN [150], VAW-GAN [151] |
| | Music generation: C-RNNGAN [152], Seq-GAN [154], ORGAN[155] |
| Video | Video applications: VGAN [168], MoCoGAN [169], DRNET [170], DVD-GAN [171] |

## 2 Background

Generative models (GM) are a rapidly advancing research area of computer vision. Generative models are the classical models for unsupervised learning where given training data $\sim p_{data}(\mathbf{x})$ from an unknown data-generating distribution generates new samples data $\sim p_{model}(\mathbf{x})$ from the same distribution. The end goal of any GM is to draw similar data samples ($p_{model}(\mathbf{x})$ from the leaned real data distribution $p_{data}(\mathbf{x})$ best explained with the help of following training objective such as:

$$\text{Generated data } p_{model}(x) \text{ want to be similar to training data } p_{data}(x)$$

**Why the generative model?**
— Realistic samples generation and handling of missing data.
— Training of GM enables the interference of latent representations can be useful as a general feature.
— Address the density estimation problem in unsupervised learning.
— It solves the problem of generating new data for training without human supervision and interventions. Generative models are essential in the perspective of modern AI.

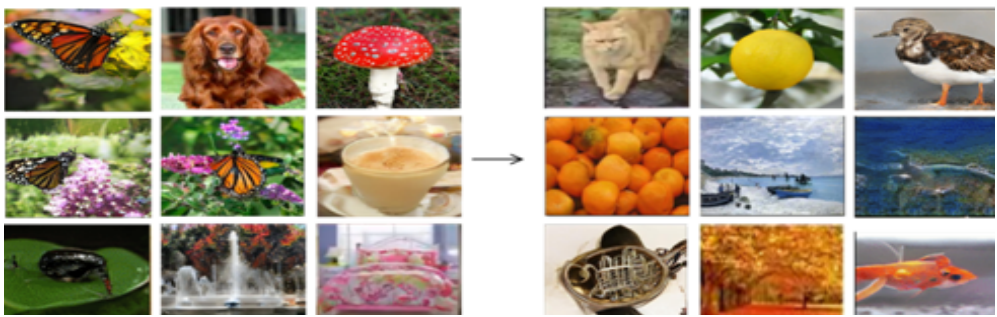

Figure 1: Some good generative models can train on examples of data and then generate more samples data. Left-side of Figure 1 shows the training examples and right-side shows the newly generated data samples — images from the ImageNet dataset [26].

## 2.1 Autoencoder (AE)

Autoencoder (AE) is the popular type of generative model that takes high dimensionality data and compresses into a small representation with the help of simple neural networks without a massive loss in data [9]. Any AE contains two types of networks: encoder and decoder networks. The encoder is a bunch of layers that takes the input data and compresses it down to small representation, which has fewer dimensions. This low or compress representation of input data is called a bottleneck. The decoder takes that bottleneck and tries to reconstruct the input data. AE calculate the reconstruction loss through per-pixel differences between encoder input and decoder output. AE objective function is given as:

$$L_{AE} = \frac{1}{n}\sum_{i=1}^{n}[x - f_{(\theta)}(g_{(\phi)}(x))]^2 \qquad (1)$$

where $g_{(\phi)}$ represents the encoder network, $f_{(\theta)}$ represents the decoder network, input data is represented by x, $\phi$, and $\theta$ represent the network parameters. A simple Euclidean distance calculates the reconstruction loss: ||input data- reconstructed data||$^2$, i.e., pixel by pixel comparison.

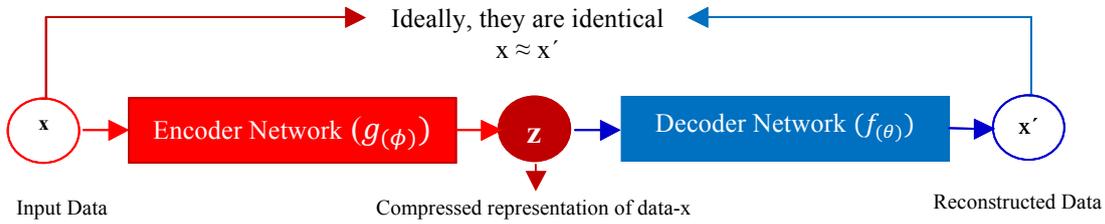

Figure 2: The simple architecture AE [9] comprises two deep networks, encoder ($g_{(\phi)}$) compresses the input data (x) through system define parameters and decoder($f_{(\theta)}$) decompress the compressed data (z).

## 2.2 Variational Autoencoder (VAE)

Variational Autoencoder (VAE) is another widely used likelihood-based generative models. It includes a probabilistic encoder network (parameterized by $\phi$), a probabilistic decoder network, or a generative network (parameterized by $\theta$) and loss functions [10]. The probabilistic encoder ($q_\phi(z|x)$) (also called latent variable generative model) embeds a data sample $x$ into discrete latent variables are denoted by z and probabilistic decoder network ($p_\theta(z|x)$) reconstructs the input sample based on the discrete latent vector z without a massive loss in input data. The cost function of the VAE is given as follows:

$$L_{VAE} = \mathbb{E}_{q_\phi(z|x)}[\log p_\theta(x|z)] - D_{KL}[(q_\phi(z|x)||p_\theta(z)] \qquad (2)$$

where $x$ represents the real data distribution, $\phi$ and $\theta$ represents the parameterize distribution for VAE probabilistic encoder-decoder structure, the first part of equation represents reconstruction loss and the second part, $D_{KL}$ stands for non- negative KL-divergence between real and the approximate posterior.

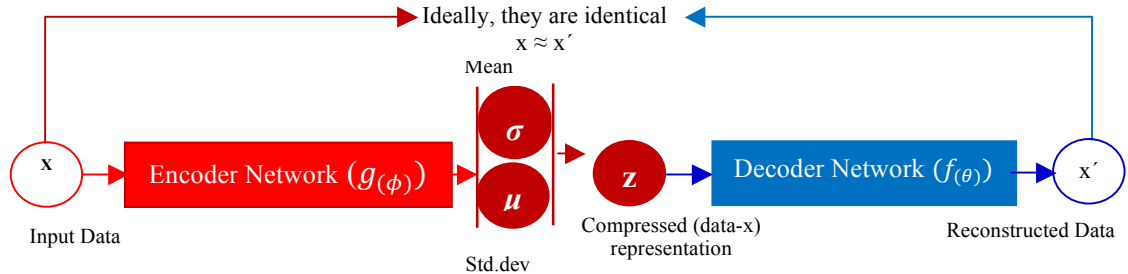

Figure 3: VAE [10] comprises a probabilistic encoder ($q_\phi(z|x)$) and a probabilistic decoder ($p_\theta(z|x)$).

## 2.3 Generative Adversarial Networks (GAN)

Generative Adversarial Networks (GAN), a robust network used for unsupervised machine learning to build a min-max game between two-player, i.e., setting up both the player (networks) with their different objectives. One player is called the generator network (*G*), and the other is called the discriminator network (*D*). 1st player (*G*) tries to fool the 2nd player (*D*) by producing very natural looking real-world images from random latent vector z, and 2nd player (*D*) gets better in-distinguishing between real and generated data. Both the networks try to optimize themselves in the best way to accomplish the individual objectives because both have their objective functions, i.e., *D* wants is to maximize its cost value, and *G* wants to minimize its cost value is given as follows:

$$L_{\text{GAN}} = V(D,G) = \mathbb{E}_{x \sim p_{\text{data}}(x)}[\log(D(x)] + \mathbb{E}_{z \sim p_z(z)}\left[\log\left(1 - D\left(G(z)\right)\right)\right] \quad (3)$$

where Equation 3 shows that there are two loss functions - log (*D*(x) for the discriminator network and log (1−*D* (*G*(z))) for the generator network and two optimizers for the generator and the discriminator since they are two different networks.

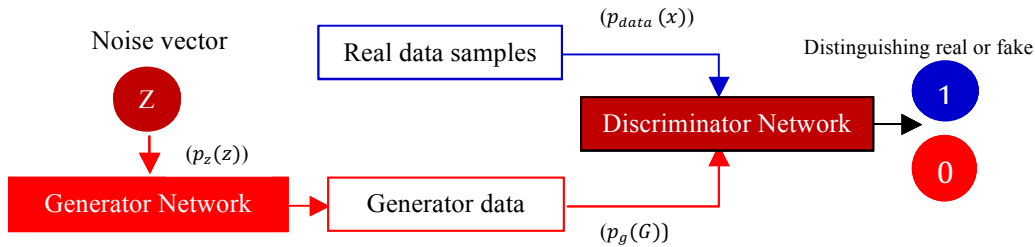

Figure 4: Design of the GAN architecture [1], where the objective of the discriminator is to maximize its cost value, i.e., log (D(x) and the generator is to minimize its cost value, i.e., log (1−D (G(z))).

## 2.4 Glow

Glow is a flow-based generative model [11] that shows the improved performance in terms of log-likelihood as compared to earlier popular generative models, such as GAN [1] and VAE [10]. The key idea behind this multi-scale architecture is a series of flow steps such as (i) Actnorm Layer-affine transformation of the activations functions through the bias-parameter and channel-scale for normalization functions, (ii) Invertible 1 × 1 convolution-fixed permutation changed with learned 1x1 convolution whose weight matrix initialized as a random rotation matrix, and (iii) Affine Coupling Layers-coupling multiple layers for numerical stability. The critical properties of the flow-based generative models are:

— Precise latent variable presumption and log-likelihood estimation:
  — VAE infer only approximately the value of the latent.
  — GAN has no encoder at all to infer the latent.
  — Glow directly infers the latent and optimizes the exact log-likelihood of the data.
— Efficient inference and efficient synthesis:
  — Autoregressive models, such as the PixelCNN [14] is inefficient on parallel hardware.
  — Glow is efficient to parallelize.
— Valuable latent space for downstream tasks:
  — Autoregressive models have unknown marginal distributions.
  — In GAN, data points cannot usually directly represented in a latent space.
  — Glow and VAE infer meaningful latent
— Significant potential for memory savings:
  — Computing gradients require a constant amount of memory.

## 2.5 Vector Quantized Variational AutoEncoder-2 (VQ-VAE-2)

Vector Quantized VAE-2 (VQ-VAE-2) [12] is an improved version of Vector Quantized VAE (VQ-VAE) [13] that generates a large size image. The key idea behind this simple feed-forward encoder and decoder architecture is using discrete latent space representation means that instead of thinking in pixels, it thinks more in terms of features that commonly appear in natural photos, which also makes the generation of these images up to 30 times quicker, and super-useful in case of larger images. This technique rapidly generates new diverse images with a size of approximately 1000 X 1000 pixels. VQ-VAE-2 is a two-stage training process. Stage-I of VQ-VAE-2 first encodes the input image into two different latent maps, one for local information, and another for global details of the input image. Stage-II of VQ-VAE-2 imposes an Autoregressive model [14] on both levels of latent space to compress the compressed image during Stage-I further. Then the simple feed-forward decoder reconstructs original size image by taking all levels of latent input maps. VQ-VAE based generative models are robust against mode collapse and lack of diversity shortcomings as compared to rivalry GAN.

## 2.6 GAN variants

This part of Section 2 discusses GAN-variants and provides a comparative analysis among them.

### 2.6.1 Conditional GAN (CGAN)

Conditional GAN (CGAN) [15] is the conditional version of GAN. These types of networks can be constructed by merely feeding the extra auxiliary information (e.g., class label), which extends the GAN into CGAN. Generator of CGAN takes the extra auxiliary information c (class label, text or images) and a latent vector z so that it generates conditional real-looking data $(G(z|c))$, and discriminator of CGAN takes the extra auxiliary information c (class label, text or images) and real data x, so that it distinguishes generator generated samples- $D(G(z|c))$ from real data x. CGAN can control the generation of data, which is impossible with the vanilla GAN. CGAN updated loss function is given as follows:

$$L_{\text{CGAN}} = \mathbb{E}_{x \sim p_{\text{data}}(x)}[\log(D(x|c)] + \mathbb{E}_{z \sim p_z(z)}\left[\log\left(1 - D\left(G(z|c)\right)\right)\right] \quad (4)$$

where the input noise variables ($p_z(z)$) and conditional variable (c) are inputs in the generator network. The real data (x) and conditional variable (c) are inputs in the discriminator network.

By comparing Equation 3 and Equation 4, it is clear that the only difference between GAN and CGAN loss lies in the additional parameter (c) in both *G* and *D*.

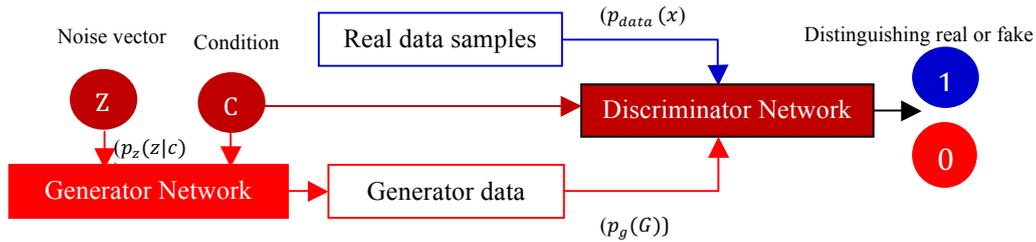

Figure 5: Illustration of the CGAN [15] architecture, where the objective of the discriminator trained on fake ($p_g(G)$) and real data ($p_{data}(x)$) is to maximize its cost value, i.e., log (D(x|c) and the generator trained on random noise vector ($p_z(z)$) is to minimize its cost value, i.e., log (1−D (G(z|c))).

### 2.6.2 Deep Convolutional GAN (DCGAN)

A new class of convolutional neural networks (CNN) [16] called Deep Convolutional GAN (DCGAN) [17]. DCGAN was the first structure that practiced de-convolutional neural networks (de-CNN) structural design that significantly stabilizes GAN training. These frameworks consist of two networks; one network works as a CNN called the generator, and the other network works as a de-CNN called discriminator. A newly proposed class of architectural constraints included in the CNN architecture is:

— Remove all levels of pooling layers with stride convolutions.
— Both *G* and *D* must use Batch Normalization (BN) [192].
— Use ReLU and Leaky-ReLU in the generator and the discriminator networks, respectively.

### 2.6.3 Laplacian GAN (LapGAN)

A sequential image generation framework Laplacian GAN (LapGAN) [18] proposed by combining the CGAN [15] model with the framework of the Laplacian pyramid (LP) [19]. LapGAN requires the multi-scale generation process in which a serious of the GAN generates particular levels of details of an image in an LP representation. The GAN at each generation step of the LP can be different. The LP was built from a Gaussian pyramid (GP) using up-sampling u (.) and down-sampling *d* (.) functions explained as:

Let $G(I) = [I_0, I_1 ... I_K]$ be the GP where $I_0 = I$ and $I_K$ are *k* repetitive applications of *d* (.) to I. Then, the coefficients $h_k$ at each level k of the LP ($L(I)$) is given as:

$$h_k = L_k(I) = G_k(I) - u(G_{k+1}(I) = I_k - u(I_{k+1})) \quad (5)$$

The rebuilding of the LP coefficients $[h_1 ... h_K]$ through backward recurrence is:

$$h_k = I_k = u(I_{k+1} + h_k) \tag{6}$$

While training LapGAN, there is a set of generative models $[G_0 ... G_K]$, each of which captures the distribution of coefficients $h_k$ for different levels of the LP. Here, while rebuilding; the generative models are used to produce $h_k$. Equation 6 is thus updated as follows:

$$\tilde{I}_k = u(\tilde{I}_{k+1}) + \tilde{h}_k = u(\tilde{I}_{k+1}) + G_k(z_k, u(\tilde{I}_{k+1})) \tag{7}$$

where at each level, a stochastic option was prepared to construct the coefficient $h_k$ using the standard procedure or $G_k$.

### 2.6.4 Information Maximizing GAN (InfoGAN)

Information Maximizing GAN (InfoGAN) [20] proposed an idea of a representation learning algorithm that can learn disentangled design in a wholly unsupervised way. InfoGAN is a completely unsupervised framework built on top of GAN and disentangles both discrete and continuous latent factors, scale to complicated datasets, and requires no more training time than GAN. The InfoGAN training objective is:

$$L_{\text{InfoGAN}} = V(D, G) - \lambda I(c; G(z, c)), \tag{8}$$

where $\lambda$ refers to a hyper-parameter, $z$ refers to un-interpretable noise, $c$ encodes the salient latent codes, and $I$ is for the mutual information /shared information.

The uniqueness of the InfoGAN compared to the standard GAN is the introduction of a regularization term (I) as it is shown in Equation 8 that captures the shared information among the interpretable variables (c) and the generator output. The Mutual Information (I) is mathematically written as:

$$I(c; G(z, c)) = Entropy(c) - Entropy(c|G(z, c)) \tag{9}$$

where the second entropy term needs admittance to the posterior (c|G (z, c)), was approximated by $D$.

Similar to InfoGAN, Semi-supervised InfoGAN (SS-InfoGAN) [21] takes the advantages of supervised and unsupervised learning via optimizing the mutual information between the un-supervised latent code and the synthesized data, SS-InfoGAN can learn the latent code representation from a smaller size unlabeled datasets more efficiently compared to the unsupervised InfoGAN.

### 2.6.5 Energy-Based GAN (EBGAN)

Energy-Based GAN (EBGAN) [22] is a variant of the GAN architecture which combined AE [9] and GAN [1] frameworks where discriminator works as an energy function refers to the low energy to actual data and high energy to fake data instead of a routine GAN probability function that determines input as actual or fake. EBGAN used two distinct losses for the training of both $G$ and $D$. When the generator of EBAGN was far from convergence, they got better in quality gradients and performance. The discriminator and the generator losses are formally given in Equation 10 and Equation 11, respectively:

$$L_D(x, z) = D(x) + [m - D(G(z))]^+ \tag{10}$$

$$L_G(z) = D(G(z)) \tag{11}$$

$L_G$, $L_D$ and $D$ are the correspondence with generator, discriminator, and reconstruction losses. These equations also show that minimizing generator loss $L_G$ concerning parameters of $G$ is similar to maximizing the $L_D$ concerning parameter $D$ has the same minimum for the positive margin m.

### 2.6.6 Wasserstein GAN (WGAN)

Wasserstein GAN (WGAN) [23] proposed a substitute loss function derived through Earth-Mover (EM) or Wasserstein distance. Not like the standard GAN cost function where discriminator works as a binary classifier function, the discriminator in WGAN used to fit the Wasserstein distance. WGAN is the most straight-forward to train using an alternate cost function that is not suffered from the vanishing gradient problem and partially remove the mode collapse obstacle to stabilize the GAN training and get better results in term of "mode collapse" problem. The Earth-Mover (EM) distance is defined as follows:

$$W(Pr, Pg) = inf_{\gamma \in \Pi(Pr, Pg)} \mathbb{E}_{(x, y) \sim \gamma}[\|x - y\|], \tag{12}$$

where $\Pi(Pr, Pg)$ refers to a collection of all mutual proportions, and the range of $\gamma(x, y)$ are $Pr$ (real data) and $Pg$ (generated data).

As $inf$ (infimum term) is intractable, so, Equation 12 can be reformulated in terms of Kantorovich-Rubinstein duality [221]:

$$W(P_r, P_\theta) = sup_{|f|_L \leq 1} \mathbb{E}_{x \sim P_r}[f(x)] - \mathbb{E}_{X \sim P_\theta}[f(x)], \tag{13}$$

where $sup$(supremum term) is the least upper bound over all the 1-Lipschitz functions in Equation 13.

Similarly, WGAN Gradient Penalty (WGAN-GP) [24] framework further extended the idea of WGAN with the introduction of a gradient penalty (GP) term on the discriminator for the implementation of the 1-Lipschitz condition. WGAN-GP would improve GAN training stability, produce high-quality samples, and has much faster convergence power than WGAN-GP. WGAN-GP updated cost function is given as:

$$L_{WGAN-GP} = \mathbb{E}_{x_g \sim \mathbb{p}_g}[D(x_g)] - \mathbb{E}_{x_r \sim \mathbb{p}_g}[D(x_r)] + \lambda \mathbb{E}_{\hat{x}_g \sim \mathbb{p}_{\hat{x}}}[(\|\nabla_{\hat{x}} D(\hat{x})\|_2 - 1)^2] \tag{14}$$

where $x_r$ is the sample data drawn from $Pr$, $x_g$ is the sample data drawn from $Pg$ and $p_{\hat{x}}$ is the uniform distribution sampled between points in $Pr$ and $Pg$.

### 2.6.7 Boundary Equilibrium GAN (BEGAN)

Boundary equilibrium GAN (BEGAN) [25] keep-up an equilibrium that manages the trade-off between variety and superiority. The main goal behind BEGAN is to change the loss function. The Wasserstein distance between reconstruction loss of actual and synthesized images gives the real loss. In BEGAN, the discriminator works during training as autoencoder balances the process optimizing of *G* and *D*. The idea of making the discriminator as an autoencoder first proposed in EBGAN [22]. BEGAN cost function is:

$$\left. \begin{array}{l} L_D(x, z) = D(x) - k_t\, D(G(z)) \text{ for } \theta_D \\ L_G(z) = D(G(z)) \text{ for } \theta_{DG} \\ L_{K+1} = k_t + \alpha\, (\gamma D(x) - D(G(z))) \text{ for k training} \end{array} \right\} \tag{15}$$

where $L_G$ represents the loss of the generator, $L_D$ represents the loss of the discriminator, $L(x)$, $L(G(z))$ represents the auto-encoder $L_1$ loss of real, fake data, and equilibrium hyper-parameter "$\gamma$" respectively.

### 2.6.8 Progressive-Growing GAN (PGGAN)

The Progressively-Growing GAN (PGGAN) [28] proposed a multi-scale based GAN architecture where both *G* and *D* start its training with low-resolution image (*e.g.*, 4×4), gradually increase the model depth by adding-up the new layers to both *G* and *D* during the training process, and end-up with the generation of large scale sharp image (*e.g.*, 1024×1024). The basic idea behind PGGAN is to grow both *G* and *D* in synchrony, i.e., starting from a low-resolution image (e.g.$4^2$), doubles the resolution of generated image (e.g.$8^2$) with the addition of new layers to the both networks, and ends-up with the generation of high-resolution image (e.g.$1024^2$) as the training progresses. The growing training approach of PGGAN enables stable learning for both the networks in big resolutions, reduced training time, and workaround GAN instability training problems.

### 2.6.9 BigGAN

Due to its outstanding, large scale, indistinguishable, and high-quality image generation capacity, Big GAN (BigGAN) [29] is one of the present best models. BigGAN-a deep learning model-can train bigger neural networks even more parameters; create a more extremely detailed image with remarkable performance. BigGAN have some essential properties such as provides exerts control over the outputs, provides interpolation phenomena between images, which means that if there are two images, it can compute the intermediate image between them and provides the best inception score (IS), i.e., the best of earlier works had an inception score (IS) around 50, but the inception score (IS) of BigGAN technique is not less than 166, which is closer to real images which would score around 233. Extension to BigGAN proposed called Bi-Directional Big GAN (BigBiGAN) [30] improves the un-conditional image generation and representation learning capacity of the model such as increased freshet inception distance (FID) and inception score (IS) accuracy score over the baseline BigGAN [29] model for un-conditional results

#### 2.6.10 Style-Based Generator Architecture for GAN (StyleGAN)

Although PGGAN [28] generates a high-quality image, its ability to control specific features of the generated image is minimal. To curb this issue, Style-Based Generator Architecture for Generative Adversarial Network (StyleGAN) [31] redesigns the architecture of the generator network, makes it possible to control the image synthesis through scale-specific amendments to the styles with-out compromising the generated image quality but increases it significantly utilizing PGGAN. Infarct, StyleGAN introduced an upgraded version of PGGAN, which only focusing on the generator network to control the co-relation between input features. StyleGAN divides the input features into three types, such as (i) coarse features-pose, hair, face, shape, (ii) medium features-facial features, eyes, and (iii) fine features-color scheme. The StyleGAN is famous for its un-conventional GAN architecture such as the use of mapping network that first transforms the input latent code into inter-mediate latent code where affine transformation then produce styles that control the layers of the synthesis network through Adaptive Instance Normalization (AdaIN) [32] that scales the normalized input with style spatial statistics, and the PGGAN [28] hat has been extremely flourishing in stabilizing large-resolution GAN training. StyleGAN2 [33] comes with various improvements to image quality, efficiency, diversity, and disentanglement, and the results are incredibly improved. StyleGAN2 simply redesigns the normalization used in the generator of StyleGAN [31], which removes the artifacts such as blob-shaped artifacts that resemble water droplets. The StyleGAN2 achieves excellent results in face image synthesis and quality than StyleGAN.

## 2.7 Analysis of GAN variants

Section 2.6 discusses different prominent versions of GAN, and now Section 2.7 provides a much-needed analysis of the GANs variants in-terms of their merits and demerits. Prior knowledge about the strengths and limitation of the model will be important before applying it successfully for a practical application.

CGAN [15] can control the generation of the image with its conditional variable applied on *G* and *D*. CGAN based models dictate the type of data generated through a applied condition, create a general framework for different application, i.e., not application specific, and potentially develop into a huge tool for providing new image datasets. DCGAN [17] is the first convolutional neural network (CNN) [16] based GAN architecture demonstrates steady training procedure and achieved great performance in superior quality sharp images generation tasks. But, on the removal the batch normalization layer (BN) [192] from DCGAN architecture, it is inferior in quality, and there is insufficient diversity in the generated images. BEGAN [25] uses Wasserstein distance instead of JS divergence that balances both the networks (*G* and *D*) in the training, which is fast, stable, avoid overfitting, and robust to parameter changes. In addition, BEGAN method also add a new estimated convergence measure to stabilize the training and generation of human faces with highly quality.

LapGAN [18] was the main adaptation for the GAN that up-scale LR input image to HR outputs image in coarse-to-fine fashion. LAPGAN combines the conditional GAN [15] with a Laplacian pyramid representation, which can generate more photo-realistic images than original GAN [1]. Although, LapGAN performs far more better than GAN in generating photo-realistic images, however, its coarse-to-fine fashion is quite complex because the generative models at each level can be totally different from other level generative models, are trained independently, i.e., each level generative model getting no updates from others which hurts its performance. EBGAN [22] demonstrates better convergence power and the capacity to produce realistic diverse images of high-resolution by preventing the generator not to produce samples fallen in a few modes through a regularizer loss term. However, fixed margin *m* phenomena used in EBGAN make it difficult to adapt to the changing dynamics of the discriminator and generator, which hurts the performance of the discriminator network in-reconstructing the real samples because the energy value of the generated example vary near the margin *m*. WGAN [23] use Wasserstein or Earth mover's distance as an alternative to JS divergence for comparing distributions, which gives better gradients and improves the training stability through solving the training problems (i.e., the vanishing gradient, and the mode collapse problems) very well. But WGAN model may still produce poor quality images and does not converge due to the use of weight clipping (i.e., weight clipping enforce a Lipschitz constraint on the critic network represents the discriminator network), which can lead to unwanted performance. In-addition, WGAN based models have limited capability to model complex functions.

InfoGAN [20] is another CGAN [15] based model makes the image generation procedure more controllable, and the outcome can be more interpreted through the induction of mutual information. However, InfoGANs are used preferably if datasets are not that complex such as ImageNet because in-case of complex dataset, it gives inferior quality results. The Progressive-Growing GAN (PGGAN) [28] grows progressively, have been extremely flourishing for improving quality, increasing stability and variation. PGGANs have additional benefits such as early layers converge quickly, only a few layers at a time to train from scratch and the training time reduce significantly. Although the performance of PGGAN is good, it is still not satisfied on mode collapse problem, i.e., the generator of PGGAN generates similar samples due to un-balance training of the both the networks ($G$ and $D$). BigGAN [29], one of the current best models due to its outstanding, large scale, in-distinguishable and high-quality image generation capacity. The performance of BigGAN is outstanding in large and high-fidelity diverse image generation, however, with random sampling, the diversity of the generated image is much lower than a real image of the same size and also the model has limited data augmentation ability on large-scale datasets such as ImageNet. The StyleGAN [31] improves the ability of GAN to have reasonable control over the generated image instead of focusing on generating more realistic-looking images, but it also has some characteristic artifacts such as blob-shaped artifacts that resemble water droplets due to instance layer normalization and phase artifact due to progressive growing phenomena.

## 3 Applications

GANs are an exceptionally amazing generative model in generating realistic-looking samples after the models have trained on some data. These advantages lead GAN to be applied in different fields of computer vision (CV) and artificial intelligence (AI). Here, we discuss various GAN applications in various domains, such as image, audio, and video.

### 3.1 Image domain

#### 3.1.1 Hand-written Chinese characters generation

Automatic characters generation is a difficult task even becomes more challenging such as the Chinese language has plenty of logographic characters in distinction to phonological languages of the world, such as English or French. Zi2zi framework is a CGAN based Chinese character generation model [35] directly derived and extended from the popular pix-to-pix translation model [34]. Dense-Net CycleGAN [36] generates Chinese characters that bases on unpaired training data where the source style information first encoded via the encoder, and then the extracted features move through a transfer unit. Finally, the transfer unit output through the decoder in the form of Chinese characters. Conditional Least Square GAN (LSCGAN) [37] proposed Chinese font generation model by combining a range of Chinese characters. LSCGAN generates a new Chinese character font style by fusing the existing Chinese characters font styles. GlyphGAN [38] is a style-consistent GAN based Chinese font generation model that takes two input vectors as class character vector that is associated with the character class information and style vector that is associated with style information. GlyphGAN generates diverse font characters by keeping the same style across all characters.

Furthermore, GAN have also been utilized in many other character generation applications such as a Multi-Scale Multi-Class Conditional GAN (MCMS-CGAN) [39] for the generation of multi-class hand-written realistic text data, Handwritten GAN (HW-GAN) [40] for synthesizing hand-written stroke data, and a Semi-Supervised GAN (SS-GAN) [41] for the generation of Bangla handwritten characters.

#### 3.1.2 Anime character generation

The creation of animated characters in computer games is expensive. Cost-effective anime-character generation is possible that requires a reduced amount of artistic skills with GAN. Several efforts have already made for the generation of anime characters such as Chainer-DCGAN [42] and Illustration-Style Reproduction DCGAN [43]. However, they fail to generate high-quality results and often produce blurred anime characters. The automatic anime character with GAN [44] automatically generates anime-characters' faces without compromising the quality. They used a Deep Regret Analytic GAN (DRAGAN) [45] as the basis of their GAN model, which enhanced stability and modeling performance as compared to other GAN types. Recently, Progressive Structure-Conditional GAN (PS-CGAN) [46] generates high resolution (e.g.1024 X 1024), and full-body anime characters with particular sequences of poses.

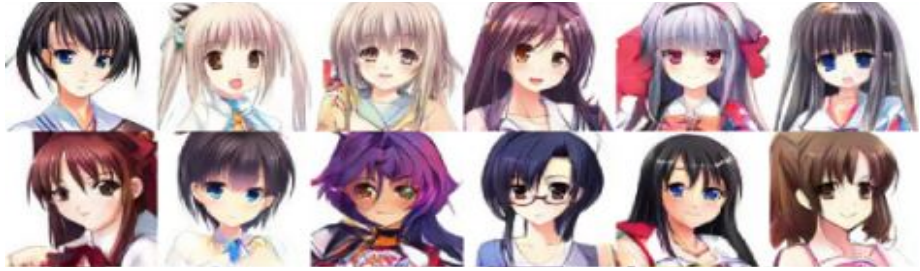

Figure 6: Examples of GAN generated anime characters faces. Images from [44].

### 3.1.3 Image blending

The blending of images together shows a dominant performance in several computer vision tasks, for example, modification of visual communication or automatic photo editing. Gaussian Poisson GAN (GP-GAN) [47] proposes an image blending method for high-resolution (HR) input images by combining CGAN [15] with traditional methods of image blending [48]. GP-GAN was the first framework that used GAN in the blending of images which have very HR. They proposed the Gaussian Poisson equation [49] to produce an HR blending image. GP-GAN fuses the information through the optimizing of [49] to produce well-blended high-resolution images, whereas preserving the HR information.

Recently, Geometrically and Color Consistent GAN (GCC-GAN) [50] combines the foreground and the background of two images seamlessly from different sources. GCC-GAN framework comprised with four sub-networks: (i) Transformation Network (TN) that generate a real-looking composite image by considering the geometric and color consistency, (ii) Refinement Network (RN) that sharps the boundary-edges of TN generated compound image, (iii) Discriminator Network (DN) that distinguishes the newly generated composite image from real image and (iv) Segmentation Network (SN) that learns to split the foreground and the background in the generated composite image.

### 3.1.4 Image in-painting

Refilling the missing pixels of an image is known as image in-painting. It is an advance reconstruction technique in the photo and video editing applications. Exemplar GAN (Ex-GAN) [51] uses exemplar-image information as a reference image to produce a high-quality result. They assumed this extra conditional information as available at an inference of time and corresponded directly to specific characteristics of the item of importance. On the other hand, the Contextual attention model [52] undoubtedly increases the efficiency of the in-painting technique by learning feature representations to match and attend the related background patches. However, current network solutions still induce undesired artifacts and noise to the repaired regions. PGGAN [53] proposed an idea by combining the global GAN [54] with a patch-wise GAN approach [55] to design a new discriminator network. PGGAN has good results than the previous in-painting methods by aggregating local and global information.

### 3.1.5 Face aging

Face aging anticipates that how a person looks in the future while age regression analyzes that how a person looks in the past subject to his present face and other relevant information. Age CGAN (Age-CGAN) [56] proposed for face aging to anticipate one's future looks and past looks, and Conditional Adversarial Autoencoder (CAAE) [57] simulates facial muscle sagging caused by aging. Similarly, Identity-Preserved CGAN (IPC-CGAN) [58] imposes an age classification term to generate photo-realistic faces and an identity-preserved term into age synthesis. Wavelet-Domain Global and Local Consistent Age GAN (Wavelet-GLCAGAN) [59] for age progression and regression. Similarly, Wavelet-GAN based face aging method [60] make sure the steadiness of features through inputting the facial feature vector along-with face image as inputs to both $G$ and $D$.

### 3.1.6 Text-to-image synthesis

Generating high-quality photo-realistic images from the text has tremendous applications in computer vision, including photo-editing, computer-aided design, and image synthesis. The Attentional GAN (AttnGAN) [61] introduced text-to-image translation model. AttnGAN is a multi-stage generation model that progressively generates LR to HR image with m-number of generators. AttnGAN consists of the attentional generative network that first draws sub-regions by applying conditions on given word descriptions that are mainly related to those regions and multimodal similarity model that calculate the resemblance between the image generated and the conditioned sentence. Stacked GAN (StackGAN) [62] proposed for text-to-image translation framework by applying condition on given text descriptions via CGAN [15]. Two GAN are set consecutively in StackGAN, where Stage-I GAN produces an image of LR on the specified text information, and Stage-II GAN enhances the resolution of the generated image.

Similarly, Auxiliary Classifier GAN (AC-GAN) [63] and Text ACGAN (TAC-GAN) [64] are text to image generation frameworks, where c (class label) concatenates with z (latent vector) [63], and z (latent vector) concatenates with the text embedding, not on class labels [64]. Semantic Image Synthesis GAN (SIS-GAN) [126] proposed for text-to-image manipulation by employing text descriptions as conditions. The generator in SIS-GAN follows the encoder-decoder structure where the encoder first encodes the input image into feature representation and then concatenates the feature representation with a text description. The decoder decodes the joint representations to a synthesized image. The discriminator determines whether the input image is realistic and whether the input image fits the definitions in the input text. SIS-GAN has quite distinct and useful applications, like image manipulation system.

### 3.1.7 Human pose synthesis

Pose transfer has many applications in computer vision; for instance, the character body in the movie-making can be manipulated/estimated into the desired pose or the generation of the training data for the human pose. Pose Guided Person Generation Networks (PG$^2$) [65] use the pose information explicitly. Two GAN are set consecutively in PG$^2$ where the Stage-I of PG$^2$ is also called Pose Integration Network (PIN) generates initial pose transferred results by capturing the global formation of a person body, and the Stage-II of PG$^2$ is also called Image Refined Network (IRN) generates sharper realistic images to bring Stage-I results closer to target image via adversarial training. Inspired by Ma et al. [65], a single-stage Deformable GAN [66] introduced to generate the human target image. Deformable GAN generates a person's image conditioned on appearance and poses information. Deformable GAN used deformable skip connections to transfer correct details, and nearest-neighbor loss to eliminate a different kind of deformation between the synthesized and the ground-truth images.

### 3.1.8 Stenographic applications

Steganography is the collection of methods utilizes to hide the secret information, e.g., a document, an image, or a video, within non-secret information. Stenographic GAN (S-GAN) [67] and Secure SGAN (SS-GAN) [68] hide the secret information. The only difference between SGAN and SSGAN is that SSGAN is more reasonable for inserting messages with the arbitrary key compared to the SGAN, which is more reasonable for inserting messages with the same key, but when using the arbitrary key, SGAN becomes unreliable. In recent years, Stegano-GAN [69] is the most popular approach uses to hide arbitrary size binary data in images. Stegano-GAN consists with three networks: (i) an Encoder Network that takes an input image and secret message of random size generates the new steganographic image, (ii) a Decoder Network that attempts to recover the embedded data, and (iii) a Critic Network that evaluates the quality of the original input image and newly produces steganographic image.

### 3.1.9 Image manipulation applications

In interactive manipulation, manipulates an image according to user goals. Tasks in applications of image editing contain to alter in color property or nature. Interactive GAN (IGAN) [70] takes the user to sketch as input and produces the most similar realistic image. IGAN can automatically adjust the output while keeping users' edits as real as promising. The Text-Adaptive GAN (TAGAN) [71] is another popular type of image manipulation application. TAGAN varies the specific visual attributes, like the color or texture of the input image, according to the applied condition on given word-text descriptions. In TAGAN, the discriminator creates word-level local discriminators according to applied word-text-description where each word-level local discriminator is attached to an explicit kind of input image visual attributes, and the generator only modify the explicit attributes of the input image according to discriminator coarse training feedback while preserving the other text-irrelevant contents, like layout and the pose of the image.

Dissection GAN [234] framework introduce a new technique to visualize the inner workings of a neural generator network and enable the users to not only edit the images, a new image can be added to existing images or existence image can be removed from existence images according to his wish without any artistic skills. Similarly, through interactive applications, users can make photo modifications as real as possible, which is one of the reasons why studying GAN that create different kinds of high-quality realistic images. Introspective Adversarial Networks (IAN) [72] is introduced by combining VAE [10] and GAN [1]. IAN is a custom-made user photo editing application called Neural Photo Editor. In Neural

Photo Editor, a user can make any adjustment on the given photo with the help of rough brush striking. Another high eminence facial characteristic editing framework referred to as Attribute GAN (AttGAN) [73] uses encoder-decoder structural design and takes the facial feature information as a component of the latent vector demonstration that is a missing-gradient in previous facial editing applications.

### 3.1.10 Visual saliency prediction

Visual Saliency describes the region of interest in an image or video that attracts human attention. Extensive studies have been made on saliency detection to get accurate results. Saliency Prediction GAN (Sal-GAN) [74] proposed two networks based on saliency prediction strategy. Sal-GAN's generator predicts saliency-maps from input image's raw-pixels, and the discriminator takes the output of the first one to discriminate a saliency map into predicted or ground truth while updating the generator parameters. Capsule-CGAN based saliency detection (Sal-CapsuleCGAN) [75] method integrates the popular capsule blocks [235] into both $G$ and $D$ instead of traditional U-Net-like structural design. The generator predicts saliency-maps not able to be distinguished from the real saliency maps, and the discriminator distinguishes the predictable output from the synthetic ones. The major draw-back of the mentioned salient object detection method is that they are unable to detect salient objects in a noisy scene. So lately, GAN based De-noising Saliency Prediction GAN (DSAL-GAN) [76] detects the salient object in a noisy image. DSAL-GAN comprised of two GAN, where the first GAN de-noise the noisy image, and the second GAN detects the salient object in a de-noised image.

### 3.1.11 Object detection

Small objects detection (SOD) is a demanding task due to the small size and noisy version. The task becomes even more complicated when their appearance is invisible with other visible objects in the scene. Segmentor GAN (SeGAN) [77] detects objects by combining the object segmentation and object generation techniques. The segmentor network of SeGAN takes an image and visible regions as inputs and generates the mask of the entire occluded object. Both $G$ and $D$ trained in an adversarial way for the generation of an object image that has invisible regions. For SOD, Perceptual-GAN (PGAN) [78] generates ultra-resolved descriptions of small objects for better detection by narrowing the differences between small and large objects and Multi-Task GAN (MTGAN) [79] uses a super-resolution network (SRN) to up-scale the small-scale distorted image into the large-scale clear image for better detection. Recently, the GAN-based Detection of Objects (GAN-DO) [80] method learns an adversarial objective for object detection through training. GAN-DO takes the low-quality image as input for accurate object detection as compared to previous object detection methods takes a high-quality image as input.

### 3.1.12 3D image synthesis

Advances in 3-dimensional (3D) volumetric convolution networks have led to the application of GAN to generate 3D objects. 3D-GAN [81] expands the system from 2D-GAN to 3D-GAN. Similarly; Projective GAN (Pr-GAN) [82] inferred 3D objects from multiple 2D views of an object. They used a 3D volumetric convolution network to generate 3D shapes in the generator, where they used 2D projected image as an input of discriminator to match synthesized 3D objects (fake object) with the 2D view. They also introduced a 3D-VAE-GAN as an expansion to 3D-GAN by including an image encoder network, which outputs the latent vector representation by taking a 2D image as input.

Furthermore, GANs have utilized in many other 3D image generation applications such as 3D-CGAN [83] uses paired training data for the generation of 3D-samples in different and controllable rotation angles, Tree-GAN [84] for 3D point cloud generation of multi-class applications, and 3D GAN (3DGAN) [85] that produces 3D structure from a latent vector using the advances in modern convolution networks.

### 3.1.13 Medical applications

GAN based medical image applications is a well-accepted active area of research. Segmentation GAN (SegAN) [86] proposed a medical image segmentation method where a critic network is used instead of the discriminator and the segmentor network instead of the generator. They use the critic because regular discriminator's single value result (1 or 0) may not be sufficient for dense, pixel-level segmentation of medical images, and a fully CNN based segmentator as a generator generates segmentation label maps. Medical GAN (Med-GAN) [87] generates realistic synthetic electronic health records (EHRs) through a combination of an AE [9] and GAN [1]. Med-GAN takes EHRs dataset as input and generates high-dimensional discrete variables. Med-GAN generated EHRs play a vital role in expert medical opinion.

Furthermore, GANs have also been utilized in many other medical applications such as drug discovery GAN (Chem-GAN) [88] for the determination of internal chemical diversity of drugs, feedback GAN (FB-GAN) [89] for generating and designing deoxyribonucleic acid (DNA), medical image processing [90] for the measurements of brain activity (e.g., MRI, fMRI), missing teeth restorations [91], on-line doctor recommendation framework (DR-GAN) [92] for finding a suitable doctor for the treatment of their health based on on-line questions and answers, and Med-GAN [93] for medical I2I translation.

### 3.1.14 Facial makeup transfer

In facial makeup transfer, the facial makeup style of one face is transfer to a non-makeup face without losing the identity of the non-makeup face, and makeup removal system does the reverse. Beauty GAN (B-GAN) [94] transfers the makeup style from makeup face to non-makeup face. B-GAN contains dual input/output GAN, without any pre/post-processing steps, takes the source and reference face as inputs, and produced new after-makeup face. Paired-CycleGAN [95] introduced a makeup transfer and removal system where the first network transfers the makeup style and other doe's inverse with the end goal that the output of their successive applications to an input photo match the input. Inspired by [94], Disentangled Makeup Transfer GAN (DMT-GAN) [96] model proposed for facial makeup transformation where the input image first decomposed into two independents parts, the makeup-style, and personal identity parts, and then two encoders-one decoder mechanism is applied (one encoder for personal identity, second encoder for makeup style, and the single decoder for the reconstruction of the original image). The discriminator in DMT-GAN is applied to make a distinction between real and generated images.

### 3.1.15 Facial landmark detection

The objective of facial landmark detection is to identify human faces distinguishable vital points. Landmark detection has various applications like facial re-enactment, head pose estimation, face recognition, and 3D face reconstruction. Style-Aggregated Networks (SAN) [97] proposed an idea for facial landmark detection. SAN is a two-stage training process where Stage-I of SAN generator generates a pool of style-aggregated face image (mean face) to cover the significant variance in image style, and Stage-II of SAN generator take the mean and original face to exploit a different kind of useful information. Then a cascade technique was used to produce predictions of the heat map. Another GAN based face landmark locations framework [98] uses retrieved uniform face landmarks as an attribute for SVM classifier for correct classification between real human face images and synthesized fake human face images. This method tries to determine that facial landmarks of the generated human faces different from real face due to the lack of global constraints.

### 3.1.16 Image super-resolution

The purpose of image super-resolution (SR) is to up-scale the images of low-resolution (LR) because images with high-resolution (HR) have better visual qualities. Super-Resolution GAN (SRGAN) [99] takes the LR input image and induces the HR output image through 4x up-scaling factors. Deep Enhanced SRGAN (ESRGAN) [100] further improves the visual quality result of SRGAN model by making three significant changes to the overall structure of SRGAN model such as the use of Residual-in-Residual Dense Block (RRDB) that has a superior learning power and easier to train, Relativistic GAN (RGAN) [203] improves the classification capacity of discriminator and enhanced perceptual loss through VGG-19 [230].

Furthermore, Super Resolution Dual-GAN (SRDGAN) [101] is proposed that learns to solve the noise before the super-resolution (SR) problem for the generation of the noise-free high-resolution images. Deep Tensor GAN (TGAN) [102] model cascades the tensor structures, i.e., a brief illustration of the input image with DCGAN [17] architecture for the generation of a high-quality large-size realistic image.

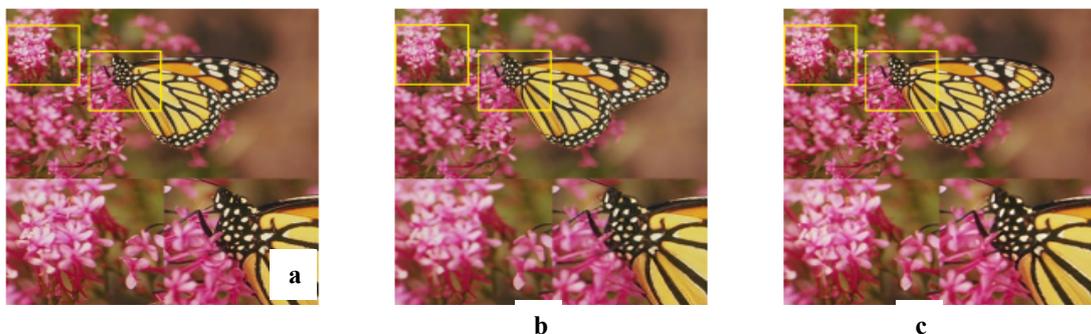

Figure 7: (a) Ground truth, (b) SRGAN result, and (c) ESRGAN result are shown in the figure, which shows that ESRGAN delivers better visual performance than SRGAN. Images from [100].

### 3.1.17 Texture synthesis

Texture synthesis is a traditional and exemplary issue in the image domain to generate a new texture that is indistinguishable from the real texture. Markovian GAN (MGAN) [103] is a texture synthesis technique based on GAN [1] with the vital dissimilarity of just working on neural patches despite full image. MGAN has a low computational cost with real-time achievement for the synthesis of neural texture compared to the primary texture synthesis method [104]. Further, Spatial GAN (S-GAN) [105] maps a spatial tensor of an image was the first edition of GAN in texture synthesis with completely unsupervised wisdom, and well appropriate for arbitrary large size texture synthesis. Similarly, Periodic Spatial GAN (PSGAN) [106] further extended the SGAN model that can synthesize different, cyclic, diverse, high-resolution, and complex datasets of texture.

### 3.1.18 Sketch synthesis

Sketch representations are a constructive way to draw what users need. Texture GAN (TGAN) [107] method proposed for sketch synthesis that directly derived and extended from the popular Scribbler method [108] with the new authority over the textures of the object. The generator of TGAN takes a texture image, a sketch image, and a color image as inputs, and generates a new image as output. Similar to TGAN, [109] introduced a CGAN based auto-painter method that automatically generates compatible colors for cartoon images. Their framework converts sketches to cartoon images by adopting the network architecture of pix2pix [34]. Sketchy-GAN [110] is another simplified sketch synthesis method that takes the sketch as input, and the output is a realistic image with the same pose from 50 diverse classes. Most recently, Composition-Aided GAN (CA-GAN) [111] takes a face mage (photo/sketch), and the related face masks as inputs and generates a new face image (photo/sketch) as output. The output of CAGAN is further refined to get sharp results through Stacked Composition-Aided GAN (SCA-GAN).

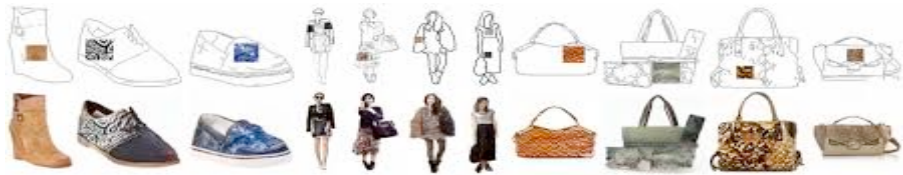

Figure 8: TGAN generated samples in the second row from first-row input sketches. TGAN can produce new examples of usual things from hand-drawn sketches and texture patches. Figure from [107]

### 3.1.19 Image-to image translation (I2I)

The purpose of the image-to-image (I2I) translation methods is to renovate the visual illustration of an image with the new visual representation. The GAN based Pixel-to-Pixel (pix2pix) [34] translation method adopts a supervised learning technique for I2I translation. The generator of pix2pix translates the source image into the target image based on the condition applied, and the discriminator makes it confirm whether the used condition is meet-up by considering the pixel-wise loss. Similarly, Perceptual Adversarial Networks (PAN) [112] introduced another method for I2I where the perceptual loss is minimized instead of minimizing the pixel loss. Cycle-GAN [113], and Disco-GAN [114] introduced I2I methods for unpaired training data. They have two different GAN coupled together; one translates style from one domain to another, and another doe's the inverse. Similar to previous unpaired image translation methods, Wasserstein distance-Dual GAN (WDGAN) [115] method used Wasserstein distance instead of a sigmoid cross entropy generative adversarial loss in addition to cyclic reconstruction loss.

Star-GAN [116] is another unsupervised I2I method that uses a single model for multi-domains I2I translation. Star-GAN has promising results on expression synthesis and facial attribute transfer tasks. Unsupervised I2I Translation (UNIT) [117] combines GAN [1] for the generation of the corresponding image in two domains through shared-latent space constraints and VAE [10] for an edited image with the input image in the relevant fields. The limit of the UNIT is that they fail to generate different results. To deal with this curb, Multimodal UNIT (MUNIT) [118] generates different results from a given basis domain image. The MUNIT first decomposed the latent space of image into style and content codes and then recombined content code in the target style space with a random style code. Diverse I2I Translation via Disentangled Representations Image-to-Image (DRIT) [119] model is a multi-model generation framework that can generate different results with un-paired training data. DRIT purpose-specific disentangled representation, which enables the efficient generation of multi-modal outputs. Likewise,

DRIT++ [120] expands the DRIT by 1) integrating sample diversity mode-seeking maximum likelihood, and 2) generalizing the two-domain structure to tackle multi-domain I2I translation issues.

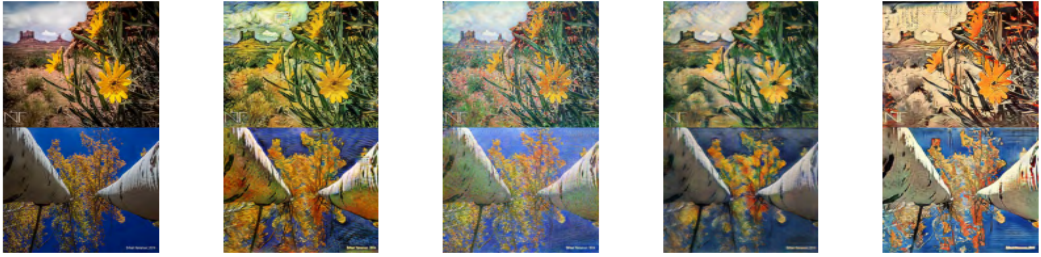

Figure 9: Some experimental results of Cycle-GAN. Images from [113]

### 3.1.20 Face frontal view generation

The enormous pose disparity among face images is a significant obstacle in the computer-based face recognition system. For automatic face recognition, frontal face activity performed to change a face image from various perspectives to frontal perspectives for better outcomes. Disentangled Representation Learning GAN (DR-GAN) [121] proposed a face frontal view generation model for the face recognition system. The generator of DR-GAN uses the encoder-decoder structural design, where encoder represents an identity attribute, and the decoder generates an image by following the posse and the encoded identity information. The discriminator of DR-GAN performs identity and poses classifications tasks. Two Pathway GAN (TP-GAN) [122] and Face Frontalization GAN (FF-GAN) [123] are others face frontal view generation models. TP-GAN generates identity preserving image after analysis of the facial-appearance in detail. FF-GAN blend 3DMM [124] into the GAN formation to give appearance and shape priors for quick convergence with a small dataset. Recently, GAN based Face-Transformation with Key Points Alignment (FT-GAN) [125] proposed for front face alteration. FT-GAN is a two-task learning process where the first task uses pixel transformation phenomena through CyleGAN architecture for frontal face synthesis, and the second task uses crucial point alignment for refining the transformed face.

### 3.1.21 Others image domain applications

There is the numbers of GAN models that have also been used in many other image domains applications, such as Cycle-Dehaze [127], Disentangled-Dehazing [128] and Cycle-Defog2Refog [129] for image de-hazing; Alpha-GAN [130] for image matting; Remote-Sensing GAN (PS-GAN) [131] for image fusions; Task-Oriented GAN (TO-GAN) [132] for image classification; Pixel-DTGAN [133] for cloth image; Gene-GAN [134] and Gaussian Poisson GAN (GP-GAN) [47] for object transfiguration; Semi-Supervised (SS-GAN) [135] for semantic segmentation; Face-mask-CGAN [136] for portrait editing; Vital [137] and Sample-Level GAN (S-GAN) [138] for object tracking; Semi-Latent GAN (SL-GAN) [139], Semantically Decomposing GAN (SD-GAN) [140], Wav2Pix [141], GAN-Fit [142], Deforming Autoencoders (DAE) [143], Dual Variational Generator (DVG) [144] and 3D Aided Duet GAN (3D-AD-GAN) [145] for face image generation; Pose Normalized GAN (PN- GAN) [146] and Identity Preserving GAN (IP-GAN) [147] for person re-identification; and Deep Convolutional GAN (DC-GAN) [148] and U-nets & GAN [149] for de-occlusion in image processing and computer vision.

## 3.2 Audio domain

### 3.2.1 Language and Speech synthesis

GAN has excellent success in synthesizing data in music generation, dialogue systems, and machine translation. Ranker GAN (Rank-GAN) [150] proposes a novel high-quality language (sentence) generation method by substituting the discriminator with a ranker network and achieved exceptional performances. The generator tries that its generated sentence is so realistic that it will be rank higher than real sentence while the ranker network calculates the ranking score of real sentence higher than generated.

A voice conversion (VC) system converts the source voice to a target voice without changing the linguistic contents was proposed called Variational Auto-encoding WGAN (VAE-WGAN) [151] combined VAE [10] and WGAN [23]. VAE-WGAN framework improves the target results with a realistic spectral shape. In VAW-GAN, the encoder provides the source voice with a phonetic substance, and the decoder integrates the transformed target voice with the information provided by a target speaker.

#### 3.2.2 Music generation

GAN has also utilized for music generation such as Continuous Recurrent Neural Network GAN (C-RNN-GAN) [152] generates continuous sequential data. C-RNN-GAN model both the networks as an RNN with LSTM [153], precisely extracting the full sequences of music. They have bad results because the distinct value of the music elements has no consideration. To have better outcomes, Sequence GAN (Seq-GAN) [154] and Object Reinforced GAN (OR-GAN) [155] consider the sequence generation method as a chronological decision-making procedure by considering the distinct property of the generated music.

#### 3.2.3 Others sequential domain applications

Furthermore, GANs have also been broadly utilized in many other natural language processing, music, speech, and voice applications, such as Utility-GAN [156] and Multi-Scale Matching GAN (MSM-GAN) [157] for question-answer selection, Resume-GAN [158] for talent-job fit, Natural Language GAN (NLG) [159] and Fake-GAN [160] for review detection and generation, Information Retrieval GAN (IR-GAN) [161] and Personalized Search GAN (PS-GAN) [162] for information retrieval, Adversarial Reward Learning (AREL) [163] for visual storytelling, Speech Enhancement GAN (SE-GAN) [164] for speech enhancement, Frequency-Domain Speech Enhancement GAN (FSEGAN) [165] for speech recognition, Caption-GAN [166] for image caption, andImage-to-Poetry GAN (I2P-GAN) [167] for poetry generation.

### 3.3 Video domain

#### 3.3.1 Video applications

Here, we discuss video generation GAN. Commonly, the video is mostly a combination of stationary background and moving objects, where predicting the object motions is a core issue in computer vision. GAN based video generation (VGAN) [168] technique decomposes the video frame into the content and motion parts. VGAN contains two generators, one for moving foreground and second for static background, separately. Motion and Content GAN (MoCo-GAN) [169] introduce a video generation framework. The proposed framework generates the video in an unsupervised way by decomposing the video into motion and content part of the latent space. Recently, the Disentangled Representation Net (DRNET) [170] approach learns disentangled image representations from the video. DRNET architecture consists of two encoder networks that produce distinct attribute re-presentations of content and pose, and one decoder network that predicts the future frames after receiving the concatenated results from encoders. Dual Video Discriminator GAN (DVDGAN) [171] generates high-resolution videos of comparatively high reliability. DVDGAN built upon the BigGAN [29]consists of one *G* and two *D*.

#### 3.3.2 Others video domain applications

GANs models have also been utilized in many other video applications such as Pose-GAN (PGAN) [172] and Dual-Motion GAN (DMGAN) [173] for future frame (video) prediction; Recycle-GAN (RGAN) [174] for video retargeting, and Video Question Answering via Multi-Modal CGAN (VQA-MM-CGAN) [175] for video question answering.

### 3.4 Miscellaneous GAN applications

GANs have also been used in many other domains such as malware detection [176], chess game playing [177], network pruning [178, 179], spatial representation learning [180], mobile user profiling [181], data augmentation [182, 183], heterogeneous information networks [184], privacy-preserving [185 - 188], social robot [189], cipher cracking [190], auxiliary automatic driving [236], continual learning [237], molecule development in oncology [238], GANs for finance [239-242], GANs for textile [243, 244], GANs for e-commerce [245], GANs for fluid-flow [246, 247 ], and in many others dominant areas.

## 4 GAN training

In this section, we survey several training obstacles associated with GAN training as well as several training techniques to improve GAN training for the generation of more realistic data.

### 4.1 Problems with training GAN

GANs are influential generative models but deeply hurt from un-unstable training due to several challenges associated with GAN training. Some of them reviewed in this section for detail discussion.

### 4.1.1 Nash equilibrium

Training of GAN may be considered as two deep neural networks, competing for one against the other in an adversarial way for the search of Nash-equilibrium, i.e., a state where neither the discriminator nor the generator can improve their cost unilaterally [191]. The generator and discriminator train themselves simultaneously [1] for Nash-equilibrium. On the contrary, when both *G* and *D* update their cost function independently without any coordination, it is hard to achieve Nash-equilibrium. Thus, GAN training becomes unstable. To stabilize the training of GAN, Nash-equilibrium is very important, which is best explains with the help of the following example:

For instance, suppose that there are two player, A and B, which manage the value x and y. Player A desires to minimize the value (xy) while B wants to maximize the value function:

$$V(x, y) = xy \tag{16}$$

By solving Equation 16, $\partial xV(x,y) = 0$ and $\partial yV(x,y) = 0$, we can determine that x = y = 0 has a saddle point (Nash-equilibrium).

Let's see whether we can find the Nash-equilibrium using gradient descent optimizer. For this, we update the parameter x and y based on the gradient descent of the value function *V* (Equation 16) describes through the partial differential equations system:

$$\Delta x = \alpha \frac{\partial(xy)}{\partial(x)} \tag{17}$$

$$\Delta y = -\alpha \frac{\partial(xy)}{\partial(y)} \tag{18}$$

where $\alpha$ is the learning rate. By plotting x, y, and xy against the training iterations, we realize that our solution does not converge because after every gradient update causes massive fluctuation and instability becomes poor, as shown in Figure 10.

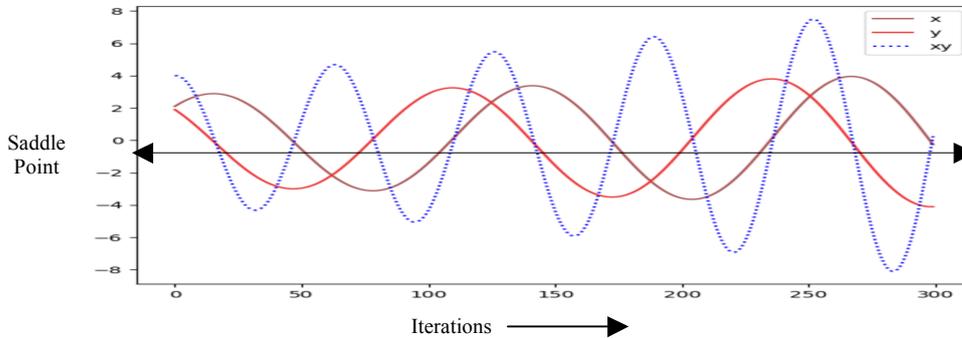

Figure 10: With more iteration, the oscillation grows-up, which causes more and more, training instability. Image from [191].

### 4.1.2 Internal covariate shift

Internal Covariate Shift (ICS) occurs when the input distribution of network activation differs as a consequence of updating parameters in previous layers [192]. When the input distribution of network changes, intermediate layers (hidden layers) try to learn to adapt to the new distribution. These learning parameters slow down the training of the model due to a change in learning rates. Due to the updated learning rates, the model required much longer training time to counter these shifts. The longer time automatically increases the training cost because the model reserved the resources in higher time.

### 4.1.3 Mode collapse

The mode collapse (MC) problem is the most crucial topic associated with GAN training, where the generator always produces an identical output. MC is a common cause of failure where a generator demonstrates low diversity amongst data or generates only specific types of real samples, which limits the usefulness of the learned GAN in many applications of computer visions and computer graphics. Mode Collapse may be of partial or complete type. Partial type of MC produces images with small diversity, and completer type (worst-case scenario) provides images of a single kind with no variety [193] [194].

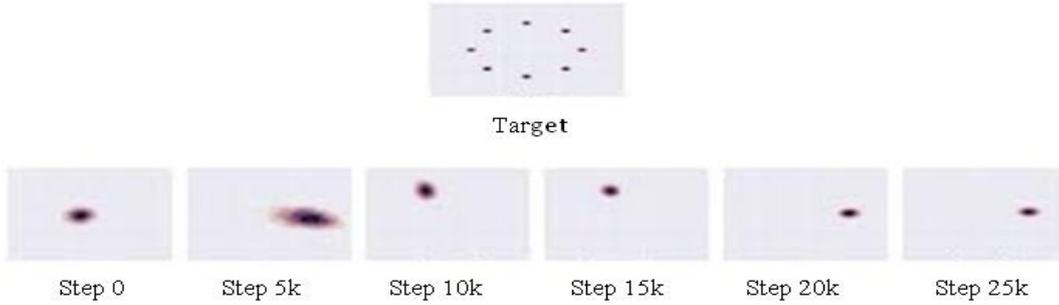

Figure 11: Visual example of mode collapse problem. Image from Unrolled GAN [197].

Figure 11 shows the problem of model collapse for a 2D toy example where the top row shows the objective distribution the model should learn, and the base row shows the number of distributions the model learned during the training. As we see in Figure 11, instead of covering all possible modes at a time, the generator covers only one mode of the goal distribution. The generator then switches to a different mode as the discriminator learns to reject the generator's selected mode.

#### 4.1.4    Vanishing gradient

GAN is measured tough to train due to the vanishing gradient (VG) problem, and the generator may fail to improve in producing good quality images. VG is due to the gradients of the generator concerning weights in earlier layers of the networks becomes very small, like vanishingly small that initial layers in the network stopped to learns [193]. The discriminator (well-trained network) rejecting the generator generated samples with confidence due to VG [1]. Optimizing the generator may be hard for the reason that the discriminator is not sharing any information which harms the learning capacity of the models.

#### 4.1.5    Lack of proper evaluation metrics

Evaluation of GAN models is an active research area regarding unstable training. Despite the GAN grand success achieved in numerous computer vision applications, it is still difficult to evaluate which method(s) is better than other methods [195] because there is no standard defined function for evaluation. Most of the papers proposed that the best GAN introduced their new evaluation techniques. Thus, there are no standard consensus parameters for fair model comparison, which hurt the GAN performance.

### 4.2    Analysis of GAN training obstacles

Section 4.1 discusses several popular training short-comings of GAN and provides an analysis to have a better understanding of the main reasoning behind these training issues with the appropriate remedies. The literature review shows that performance of GAN [1] is severe damage due to training shortcomings such as Nash- equilibrium [191] describes a particular state where both $G$ and $D$ do their best to have their best outcomes The optimization objective of GANs is to reach a Nash equilibrium stage where both the networks at the peak concerning their performance. But practically to reach this point is very difficult due to a lack of mutual communication between both the networks. Both the networks have to communicate with each other to enhance the individual performance, but when both the networks didn't communicate with each, update their cost function independently without any coordination, it is hard to achieve Nash-equilibrium. So, the lack of mutual communication is the main hurdle to reach a Nash-equilibrium point. Many methods, such as the WGAN [23], Two Time-scale Update Rule (TTUR) [198], and Least-Squares GAN (LSGAN) [220] are considered to reach the Nash- equilibrium.

Internal covariate shift (ICS) [192] refers to the continuous change in the input distribution to the network layer during system training. ICS slow-down the training process, and the network requires longer training time to converge to global minima. The main reasoning behind ICS phenomena is the continuous change in network parameter values. The network becomes deeper due to continuous changes to the network parameters values. Different normalization techniques such as Batch Normalization (BN) [192] and Weight Normalization (WN) [211] solve this problem through layer normalization strategy. Mode collapse (MC) [193][194] describes a situation in which a generator generates samples that have very little diversity. One of the main reasoning of MC is that both $G$ and $D$ get stronger and stronger with training iterations. But if one network, i.e., $G$ or $D$, becomes far more influential than the other, the learning signal to the other network becomes ineffective, and the system fails to replicate the diversity in the training dataset. The other main reasoning is that the generator is not working according to its full capacity; the result is that it does not cover all possible modes of the target distribution. Different remedies have proposed to reduce MC effects during training such as WGAN [23], Mini-Batch Discrimination (MBD) [196], Unrolled GAN (UGAN)[196], Regret Analytic GAN (DRAGAN) [45], AdaGAN [231], Mode Regularized GAN (MRGAN) [232], Multi-Agent Diverse GAN (MAD-GAN) [234] try to strengthen the generator network to expand its capacity by restricting it from optimizing for a single rigid discriminator.

GANs are considered difficult to train due to the vanishing gradient (VG) problem [193]. The main reasoning behind VG is the use of sigmoid function in standard GAN objective function for the discriminator, and powerful discriminator can be the reasoning for the VD problem. The WGAN [23], Least-Squares GAN (LSGAN) [220], Loss-Sensitive GAN (LS-GAN) [27], Relativistic (RGAN) [203], Spectral normalization GAN (SN-GAN) [216], and Batch Normalization (BN) [192] techniques are designed to avert vanishing gradients problem for the generator during GAN training which improves the training stability and convergence ability of the system.

## 4.3 Techniques to improve GAN training

Unstable training is one of the biggest problems concerning GAN due to many factors discussed in Section 4.1. Stabilize the GAN training process, and for better architecture; here, we survey several solutions in detail that are crucial for the successful implementation of the generative models.

### 4.3.1 Feature matching

This technique improves the training stability of GAN. Feature matching (FM) [196] introduces a new cost function for the generator by substituting the discriminator's output in Equation 3 to prevent over-fitting from the current discriminator. The generator generates data in the newly introduced cost function that must reflect the statistics of the actual data, and the discriminator must discriminate the statistical data of the actual data rather than explicitly maximizing the discriminator's output. FM is a practical approach to unstable GAN. The new modified cost function is defined as follows:

$$||\mathbb{E}_{x \sim p_{data}} f(x) - \mathbb{E}_{z \sim p_z(z)}(f(G(z))||_2^2 \qquad (19)$$

where $L_G$ represents the updated generator loss, $f(x)$ shows the intermediate activation of discriminator and $f(G(z))$ denote the features maps of generator generated data.

### 4.3.2 Unrolled GAN (UGAN)

As stated earlier, the model collapse is one of the critical issues when the GAN becomes unstable during training. The unrolled GAN (UGAN) [197] approach solves the problem of model collapse. In UGAN, the generator is updated by unrolling the discriminator updates steps in contrast to standard GAN, where the discriminator was first updated by keeping the generator fixed, and then the generator is updated for the updated discriminator during training. UGAN can noticeably reduce the mode dropping problem and improve the stability of GAN during training.

The optimal parameter ($\theta_D^*$) for $D$ can be defined as:

$$\theta_D^0 = \theta_D \qquad (20)$$

$$\theta_D^{k+1} = \theta_D^k + \eta^k \frac{df(\theta_G, \theta_D^k)}{d\theta_D^k} \qquad (21)$$

$$\theta_D^*(\theta_G) = \lim_{k \to \infty} \theta_D^k \qquad (22)$$

where $\eta^k$ is the learning rate, $\theta_G$ represents a parameter for generator and $\theta_D$ represents discriminator parameter networks, respectively.

By unrolling the discriminator optimization for the parametric value of $K$ steps, a new surrogate objective function for the updates of the generator network is as follows:

$$L_G = f_k(\theta_G, \theta_D) = f(\theta_G, \theta_D^k(\theta_G, \theta_D)) \qquad (23)$$

As it is shown in Equation 23, the updated lost function for the generator network can be controlled by adjusting the value of unrolling steps, i.e., the value of parameter *K*.

This new surrogate objective function (Equation 23) is used for the generator, and discriminator parameters update is defined as follows:

$$\theta_G \leftarrow \theta_G - \eta \frac{df_k(\theta_G, \theta_D)}{d\theta_G} \quad (24)$$

$$\theta_D \leftarrow \theta_D + \eta \frac{df(\theta_G, \theta_D)}{d\theta_D} \quad (25)$$

By examining the surrogate loss $f_k(\theta_G, \theta_D)$, gradient concerning the generator parameters $\theta_G$:

$$\frac{df_k(\theta_G, \theta_D)}{d\theta_G} = \frac{\partial f(\theta_G, \theta_D^k(\theta_G, \theta_D))}{\theta_G} + \frac{\partial f(\theta_G, \theta_D^k(\theta_G, \theta_D))}{\partial \theta_D^k(\theta_G, \theta_D)} \frac{d\theta_D^k(\theta_G, \theta_D)}{d\theta_G} \quad (26)$$

where the first expression represents the GAN gradient, and the second expression reflects how discriminator response to changes in the generator on the left-side in Equation 26. If the generator tendency to collapse to one mode, discriminator raises the loss for generator. Thus, the UGAN can put off the mode collapse trouble for GAN.

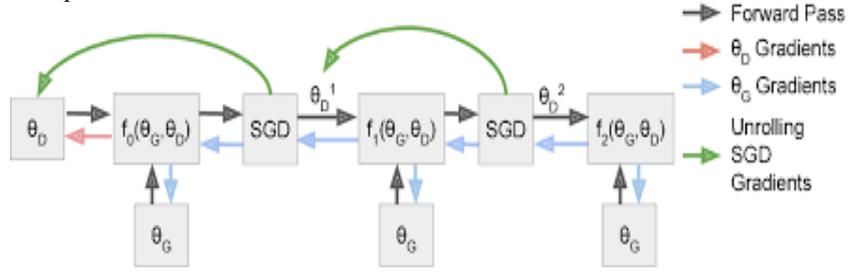

Figure 12: Example of UGAN computation for 3 unrolling steps, where the generator and the discriminator update their parameters through Equations 24 and 25, respectively. Each unrolling step k uses the gradients of $f_k$ concerning parameter ($\theta_D^k$) defined in Equation 26. Figure from [197].

#### 4.3.3 Mini-batch discrimination (MBD)

Mini-batch discrimination (MBD) [196] solves the problem of mode collapse during the GAN training. In this technique, the discriminator processes multiple data examples in mini-batches instead of processing each data example independently to avoid the mode collapse of the generator. MBD is a multi-step process that consists of the following steps to add mini-batch discrimination to the network.

Let $f(x_i) \in \mathbb{R}^A$ denotes the feature vector of the input data $x_i$ and $T \in \mathbb{R}^{A \times B \times C}$ represents the tensor vector. Mini-batch discrimination is a multi-step procedure defined in the form of the following steps:

1) Generate a matrix M with $i$ rows through multiplying the features vector ($f(x_i) \in{}^A$) by a tensor ($T^{A \times B \times C}$)

$$M_i \in \mathbb{R}^{B \times C}$$

2) Then, calculates the $L_1$-distance between the rows of the matrix ($M_i$) across samples $i \in 1, 2, \ldots, n$ and then apply the following defined negative exponential:

$$c_b(\mathbf{x}_i, \mathbf{x}_j) = \exp(-||M_{i,b} - M_{j,b}||_{L_1}) \in \mathbb{R}$$

3) Then, calculate the summation of all distance for a particular example:

$$o(\boldsymbol{x}_i)_b = \sum_{j=1}^{n}(c_b(\boldsymbol{x}_i, \boldsymbol{x}_j)) \in \mathbb{R}$$

4) Then, concatenate $o(x_i)$ with $f(x_i)$,

$$o(x_i) = [o(x_i)_1, o(x_i)_2, o(xi)_3 \ldots o(x_i)_B]\ B \in \mathbb{R}^B$$

$$o(X) \in \mathbb{R}^{n \times B} \quad (27)$$

where o(X) in Equation 27 is the result of the MBD [196] operation for the input data sample $x_i$.

#### 4.3.4 Historical averaging (HA)

Historical Averaging (HA) [196] can improve the stability of GAN during training. This approach takes the average of the parameters ($\theta[i]$) in the past at a time ( ) and adds the average value to the respective cost functions of both *G* and *D*. Historical Averaging is considering during the networks parameters updates represented mathematically such as:

$$HA = ||\theta - \frac{1}{t}\sum_{i=1}^{t}\theta[i]||^2 \qquad (28)$$

where the value of model parameters ($\theta$) at the past time ($i$) is represented by $\theta[i]$ in Equation 28.

#### 4.3.5 Two time-scale update rule (TTUR)

Two time-scales rule (TTUR) approach uses different learning rates for both *G* and *D* to speed-up the process of GAN training. It creates equilibrium among the parameters of the generator and the discriminator to update the training steps in GAN [198]. In TTUR, the generator network uses a slower learning rate, and the discriminator network uses a faster learning rate to reach Nash-equilibrium status. The discriminator network trains with a learning rate of 4 times greater than the generator network. Hence, a higher learning rate eases the problem of slow learning of the regularized discriminator. They experimentally showed that different learning rates could accelerate the GAN training process.

#### 4.3.6 Hybrid model

Every generative model has its pros, and cons like VAE [10] suffer less from the mode collapse problem but generates blurry and low-quality images. However, GAN [1] generates sharper images of high quality but suffers more from mode collapse. Recently different approaches have proposed to combine VAE and GAN into one framework called hybrid modeling. The hybrid modeling approach has vast applications in different domains of computer vision like Neural Photo Editor application (IAN) [72] achieved better-customized results by combining VAE and GAN model together.

Similarly, VAE-GAN [] architecture was proposed by sharing the decoder of a VAE with the generator of a GAN in order to supplement the reconstruction loss with a learned similarity metric, which is calculated in a feature space rather in a image space. VAE-GAN architecture can constantly produce sharp and realistically looking samples. Another hybrid model called transferred deep-convolutional generative adversarial network (tDCGAN) [] uses a DAE (Deep Auto-encoder) as a generator and DCGAN (Deep Convolution GAN) as a discriminator to detect zero-day attacks in malware- zero-day attacks are those attacks which are unidentified prior to their detection. tDCGAN model first generates fake malware, and then learns to distinguish fake malware from real malware. This junction of VAEs and GANs architectures gives better outcome and could prove to attain momentous results, particularly in provisions of preventing mode collapse.

#### 4.3.7 Self-Attention GAN (SAGAN)

In Self-Attention GAN (SAGAN), the information from a broader feature space across image regions can create samples with a fine-detailed resolution instead of information-spread in local neighborhoods [199]. SA-GAN can generate multi-classes images by coordinating the fine details of every location with distant portions. Similarly, machine interpretation models [200] accomplish best results by completely utilizing a SA-mechanism. The image transformer model [201] introduced for image generation with the addition of a SA-mechanism into an autoregressive model. [202] formalized the SA-mechanism as a non-nearby activity to demonstrate the spatial-transient conditions in a video sequence.

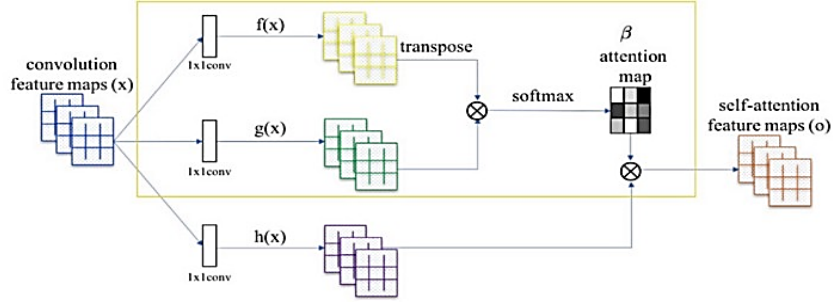

Figure 13: f(x) and g(x) represent the two feature spaces that first transform the object features from the previously hidden layer to measure the attention map ($\beta$). Matrix multiplication is indicated by $\otimes$ [199].

#### 4.3.8 Relativistic GAN (RGAN)

Relativistic GAN (RGAN) claims that the generator training should not only enhance the possibility that the synthesis sample is actual but also reduce the possibility that the actual sample is actual [203]. The discriminator in R-GAN figures out how to predict "if an image is more real than the second imager" instead of "if an image is real or fake." Experimental results show that R-GAN has training stability than vanilla GAN as shown in Figure 14. In standard formulation, new loss function for R-GAN is stated as:

$$\min_{D} \mathop{E}_{\substack{x_r \sim p_r \\ x_g \sim p_g}} (\log(\text{sigmoid}(C(x_r) - (C(x_g))))) \quad (29)$$

$$\min_{G} \mathop{E}_{\substack{x_r \sim p_r \\ x_g \sim p_g}} (\log(\text{sigmoid}(C(x_g) - (C(x_r))))) \quad (30)$$

where x is an image (real or fake), C(x) is a function that assigns a score to the input image (evaluates how much x is realistic) & σ translates the score into a probability between zero to one

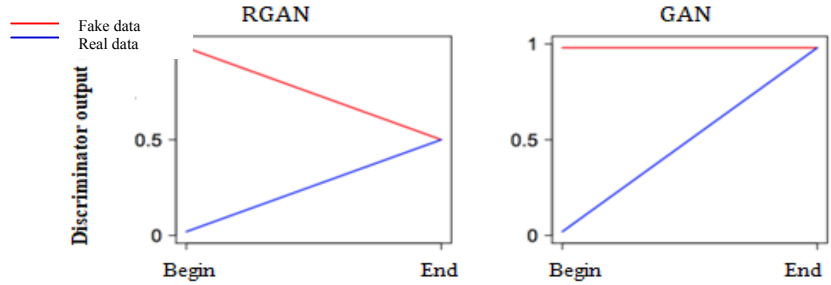

Figure 14: RGAN's discriminator do better than GAN's discriminator in distinguishing between real or fake samples. RGAN's discriminator pushes towards .5 rather than pushes towards 1 as it is happening in standard GAN. RGAN has stable and better training than standard GAN. Figure from [203].

#### 4.3.9 One-sided label smoothing

This technique provides smooth labels values to the discriminator, which significantly improves the training. The vulnerability of GAN reduced by feeding the labels with softening value, i.e., slightly less than 1, such as 0.9 for the real sample and slightly high than 0, such as 0.1 for fake samples instead of 1 and 0 [204]. Label-smoothing highly useful for the training of high-quality networks on relatively modest-sized training sets.

#### 4.3.10 Sampling GAN

Sampling GAN (S-GAN) uses a spherical linear interpolation method for data sampling instead of the linear interpolation method [205]. In S-GAN, each section bounded by two data points can be interpolated independently. They experimentally showed that the spherical linear interpolation method performs better than linear interpolation methods, as shown experimentally in Figure 15.

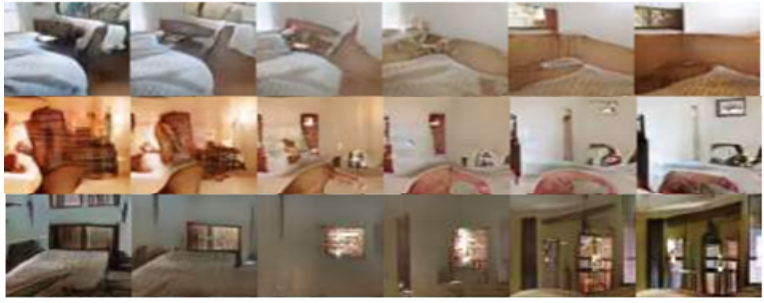

Figure 15: The top series (linear interpolation) often suffers from blurring, as shown in the middle series or another visual washout. In contrast, the bottom series (spherical interpolation) stays crisper and more visually consistent with the style of the endpoints. Images from [205].

#### 4.3.11 Proper optimizer

The use of appropriate optimization algorithm such as Root Mean Square Propagation (RMSProp) [206], Stochastic Gradient Descent (SGD) [207], Adaptive Gradient Algorithm (AdaGrad) [208] and Adaptive Moment Estimation (ADAM) [209] can dramatically improve the training stability of GAN. Experiments with different optimizers show that ADAM is the most popular optimizer among all other optimizers for severe dimensional problems and improves convergence speed and GAN training stability.

#### 4.3.12 Normalization

GANs are computationally expensive due to their more considerable training time. The training time of GAN can be reduced with the help of normalizing operations. Different normalization techniques proposed to minimize training time, which improves of stability of GAN during training. Here, we review different normalization techniques that can stabilize GAN model training.

— **Batch normalization (BN)**

Batch normalization (BN) [192] is an essential technique for getting deeper models to work without falling into mode collapse. It is a pre-processing step applied to the intermediate layers (hidden layers) that reduces the internal covariate shift and avoids mode collapse by normalizing each mini-batch of data utilizing mean ($\mu$) and variance ($\sigma$).

Let suppose a mini-batch of input data samples $\{x_1, x_2,...x_m\}$, and we compute the mean and variance such as follows:

$$\mu = \frac{1}{m}\sum_{i=1}^{m} x_i , \qquad (31)$$

$$\sigma^2 = \frac{1}{m}\sum_{i=1}^{m}(x_i - \mu)^2, \qquad (32)$$

and then replace each $x_i$ with its normalized version such as:

$$\widehat{x_i} = \frac{x_i - \mu}{\sqrt{\sqrt{\sigma^2 + \epsilon}}} \qquad (33)$$

where $\epsilon$ is the constant factor added for numerical stability in Equation 33, which shows the result of BN for data samples ($x_i$). Better normalization provides better accuracy and stable training.

— **Virtual batch normalization (VBN)**

Virtual batch normalization (VBN) [196] is an extension of the popular BN that normalizes input example by using the statistical data of many other inputs in the same mini-batch. To avoid this issue where several other inputs data statistics are being used, the VBN normalizes each example based on collected data from a batch of reference chosen on one occasion and pre-determined for the training. The reference batch normalizes statistical data only. VBN is computationally costly, so it is mostly used only in the generator .

— **Layer normalization (LN)**

Layer normalization (LN) [210] normalizes the features across each example instead of normalizing the samples across mini-batches of the training data sets. LN is advantageous over the famous BN [192] method like LN to reduce the training time considerably better than the other normalization technique.

Let suppose for input data $x_i$ of dimension $D$, we compute the mean and variance such as:

$$\mu = \frac{1}{D}\sum_{d=1}^{D} x_i^d, \tag{34}$$

$$\sigma^2 = \frac{1}{D}\sum_{d=1}^{D}(x_i^d - \mu)^2, \tag{35}$$

and then replace each component $x_i^d$ with its normalized version such as:

$$x_i^d = \frac{x_i^d - \mu}{\sqrt{\sqrt{\sigma^2 + \varepsilon}}} \tag{36}$$

Equation 36 shows the result of Layer Normalization, where the information is computed across all features ($x_i^d$), i.e., free from other examples of the training datasets.

— **Weight normalization (WN)**

Weight normalization (WN) [211] has quite distinct advantages over other normalization techniques like WN has smaller calculation cost and easy implementation as compared to other famous normalization techniques. Instead of normalizing the mini-batches directly, WN normalizes the weights of each layer to advance the optimization of the weights of neural network models. The computational cost of WN is free from batch-size of training data and has stable training accuracy across a broad span of batch size.

Let suppose; we think a neural network where the computational cost of every neuron defined as:

$$w = \emptyset(w.x + b) \tag{37}$$

where w is a k-dimensional weight vector, b is a scalar bias term, x is a k-dimensional vector of input features, ø(.) is an element-wise non-linearity, and y is the scalar output of the neuron in Equation 37.

By applying the re-parameterization trick on each weight vector w (Equation 37), speed-up the LN convergence process. Through newly defining parameters v and g, the proposed WN takes the form:

$$W = \frac{g}{||v||} v \tag{38}$$

where v is a k-dimensional vector, g is a scalar value and ||v|| is the Euclidean norm of v. By fixing v of weight vector w, we have new re-parameterization weight term, i.e., ||w|| = g (weight normalization).

Instead of optimization concerning w, Salimans, and Kingma [211] optimize concerning g and v. The corresponding derivatives take the form:

$$\nabla_g L = \frac{\nabla_w L \cdot v}{||v||}, \quad \nabla_v L = \frac{g}{||v||}\nabla_w L - \frac{g \nabla_g L}{|||v||^2} v \tag{39}$$

where $\nabla_g L$ is the gradient concerning the weights w. This equation shows the derivative for g.

$$\nabla_v L = \frac{g}{||v||} M_w \nabla_w L, \text{ with } M_w = I - \frac{ww^T}{|||w||^2} \tag{40}$$

where $\nabla_v L$ is the gradient concerning the weights v. Equation 40 shows the derivative for v. Note, Mw takes the form of a house-holder transformation on the normalized vector w/||w||$^2$.

— **Instance normalization (IN)**

Instance normalization (IN) [212] approach normalizes the features within each example instead of normalizing all the features at once across mini-batches. Quality of style transfer further improved with the use of IN. The performance of specific deep neural networks in GAN, like image generation applications, dramatically improved through IN due to batch-size limitations.

— **Group normalization (GN)**

Group normalization (GN) [213] is a simple substitute technique for the famous BN technique, which suffered from varying batch dimensions. In GN, instead of normalizing the features within each channel, they divide the channel and then normalize the features.

— **Batch-instance normalization (BIN)**

Batch-instance normalization technique (BIN) [214] proposes to normalize the redundant styles from images by binding the advantages of both BN and IN. In BIN, the IN discards the irrelevant information, and the BN preserves the critical, relevant information. So, both normalization techniques are in collaboration with BIN for better results.

Let suppose $x \in \mathbb{R}^{NxCxHxW}$ be an input mini-batch and $x_{nchw}$ denotes the elements to a specific layer, where n represents mini-batch sample index, c represents index channel, w and h specify the spatial position, respectively.

At the first stage, BN technique normalizes each example of the mini-batch through mean and variance:

$$\hat{x}^{(B)}_{nchw} = \frac{x_{nchw} - \mu_c^{(B)}}{\sqrt{\sigma_c^{2(B)} + \epsilon}}, \tag{41}$$

$$\hat{x}^{(B)}_{nchw} = \frac{1}{NHW} \sum_N \sum_H \sum_W x_{nchw}, \tag{42}$$

$$\sigma_c^{(2B)} = \frac{1}{NHW} \sum_N \sum_H \sum_W (x_{nchw} - \mu_c^{(B)})^2 \tag{43}$$

where $\hat{x}^{(B)} = \hat{x}^{(B)}_{nchw}$ is the result of BN, which discards irrelevant information.

At the second stage, IN technique normalizes each example through feature statistics of each instance:

$$\hat{x}^{(I)}_{nchw} = \frac{x_{nchw} - \mu_{nc}^{(B)}}{\sqrt{\sigma_c^{2(I)} + \epsilon}}, \tag{44}$$

$$\mu_{nc}^{(I)} = \frac{1}{HW} \sum_H \sum_W x_{nchw}, \tag{45}$$

$$\sigma_{nc}^{2(I)} = \frac{1}{HW} \sum_H \sum_W (x_{nchw} - \mu_{nc}^{(B)})^2 \tag{46}$$

where $\hat{x}^{(I)} = \hat{x}^{(I)}_{nchw}$ is the result of IN, which preserves the critical, relevant information.

BIN has to preserve important style attribute, they do this with the introduction of learnable parameters, $\rho \epsilon [0, 1]^C$:

$$Y = (\rho.\hat{x}^{(B)} + (1 - \rho).\hat{x}^{(I)}).\gamma + \beta \tag{47}$$

where the affine transformation parameters are represented by $\gamma, \beta \in \mathbb{R}^c$ and BIN output is $y \in \mathbb{R}^{N \times C \times H \times W}$, respectively:

They restrain the basics in $\rho$ with-in range of [0, 1], striking the limits at the parameter renew step,

$$\rho \leftarrow \text{clip}_{[0,1]}(\rho - \eta \Delta \rho) \tag{48}$$

where $\Delta \rho$ indicates the parameter update vector determined by the optimizer, $\eta$ is the learning rate and intuitively, $\rho$ can be interpreted as a decider, i.e., the value of $\rho$ =1 if the style (BN preserves the style) is vital to the task and the value of $\rho$ =0 if a style (the style is normalized through IN) is unnecessary.

— **Switchable normalization (SN)**

Switchable Normalization (SN) [215] introduces to tackle the different normalization techniques used for different convolutional layers within a CNN. SN technique learns to select appropriate normalizations technique such as BN [192], IN [212], and LN [210] for each normalization layer of a CNN. Different experiments have shown better performance if each normalization layer uses its normalization operation.

— **Spectral normalization (SN)**

Spectral normalization GAN (SN-GAN) [216] proposes to stabilize GAN training. It is a weight normalization technique that typically used on the discriminator where SN adjusts the weights of all the layers to 1 always performs better. This essentially ensures that the discriminator is KLipchitz continuous. EBGAN [22] further proved that SN in the generator could also stop the acceleration of parameter magnitudes and avoid unusual gradients. SN considerably reduces the computational cost and show better training behavior because other hyper-parameters don't need to be tuned. SN is defined as follows:

$$\overline{W}_{SN}(W) = \frac{W}{\sigma(W)} \quad (49)$$

where $\overline{W}$ and $\sigma(W)$ are correspondence weights on each layer for $D$ and $L_2$ matrix normalization of W.

### 4.3.13 Add instance noise

Adding the instance noise to both the real and generated data can increase the training stability of GAN [217,218]. The level of additive independent Gaussian noise can be toughening throughout the training process, and the noise allows us to carefully optimize the discriminator until convergence reached in each iteration. This is beneficial in fixing the instability and vanishing gradient issue of GAN training.

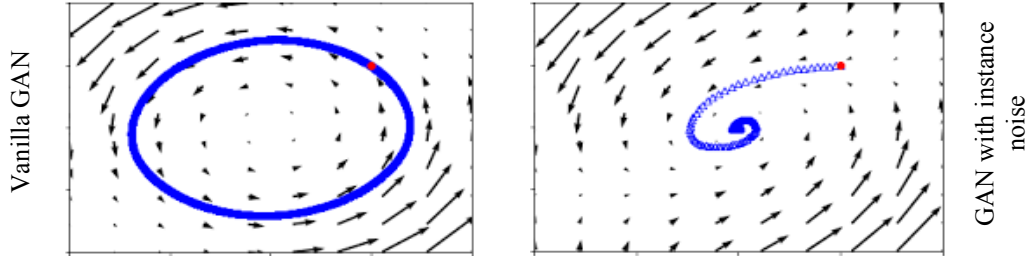

Figure 16: Computational graph shows that GAN with instance noise has better training behavior in the form of convergence power and training stability than vanilla GAN. Images from [219].

### 4.3.14 Train with labels

Adding the label as part of the latent space improves the GAN training. Conditioning model can manage the generation of samples with its conditional variables; applied to both $G$ and $D$. Conditional GAN [15] can be constructed by merely feeding the extra information in the model that controls the generation of data, which is impossible to control with vanilla GAN. Conditional GAN used in many computer-vision tasks, as Age-CGA introduced for the generation of high-quality images by considering the age condition [56], AC-GAN [63] for conditional image synthesis, and CGAN based cartoon image generation [109] from the sketch, and so on.

### 4.3.15 Alternative cost functions

Alternative cost functions prevent mode collapse and gradient vanishing. Least-square GAN (LSGAN) [220] proposed a new cost function where sigmoid cross-entropy loss is replaced with a least-square loss to overcome vanishing gradient problem. A newly introduced least square loss in LS-GAN not only classifies the actual and fake samples but also brings the fake data nearer to the real data for better architecture and real-world generation of data. The updated loss from Equation 3 is defined as follows:

$$L_{max\ D}^{LSGAN} = \max_D \frac{1}{2} E_{x \sim P_{data}}[(D(x)\text{-}b)^2] + E_{z \sim P(z)}[(D\ (G\ (z))\text{-}a)^2] \quad (50)$$

$$L_{min\ G}^{LSGAN} = \min_G \frac{1}{2} + E_{z \sim P(z)}[(D\ (G\ (z))\text{-}c)^2] \quad (52)$$

where a, b, and c are the correspondence with the label for the generated samples, the label for the real samples, and the value that $G$ desires $D$ to consider for the produced samples the discriminator to get actual values rather than getting probabilities values for real or generated values respectively.

### 4.3.16 Gradient penalty (GP)

WGAN Gradient Penalty (WGAN-GP) [24] suggests that adding a gradient penalty (GP) term in place of the weight clipping would improve the modeling performance and the stability of GAN during training. In short, the use of GP greatly enhancing training stability and reducing mode collapse of the networks. WGAN-GP has better convergence power, improved training speed, and sample quality compared to other robust GAN models by pushing the discriminator network to learn smoother decision boundaries.

### 4.3.17 Cycle-consistency loss

CycleGAN [113] based on the idea that when we translate an image from domain-A to domain-B (i.e., A→B), and then translate back from domain-B to domain-A (i.e., B→A), the result we get should be similar to original input image. During training, model minimizes the distance between original and reconstructed image. In addition, the discriminator is trained to tell whether the actual input image belong to domain-A or not. The same-thing is also done on the opposite direction, i.e., for image of domain-B. This CycleGAN architecture prevents generators from extreme hallucinations and collapsing mode by enforcing cycle consistency mechanism. So, the cycle consistency loss play a big role in-avoiding the mode collapse problem and stabilizing the training process of GAN.

## 4.4 Performance analysis

This part of Section 4 tries to describe the effects of GAN training techniques in some of its popular applications, such as the text-to-image (T2I) synthesis application (detailed in Section 3.1.16) successfully applies feature matching (FM) training technique (detailed in Section 4.3.1) to avert the mode collapse in his experiment. T2I synthesis through stack-GAN [62] has shown higher output diversity results via feature matching training techniques than previous standard GAN-based approaches. The effect of feature matching training technique in diverse outcomes can be seen in Figure 17. Similarly, the author [37] of Chinese characters generation application tries least squares loss (Least-Squares GANs (LSGANs)) [220] (detailed in Section 4.3.15) instead of cross-entropy loss in his experiments which minimize the effects of vanishing gradient problem, improves the training stability and convergence ability of the system.

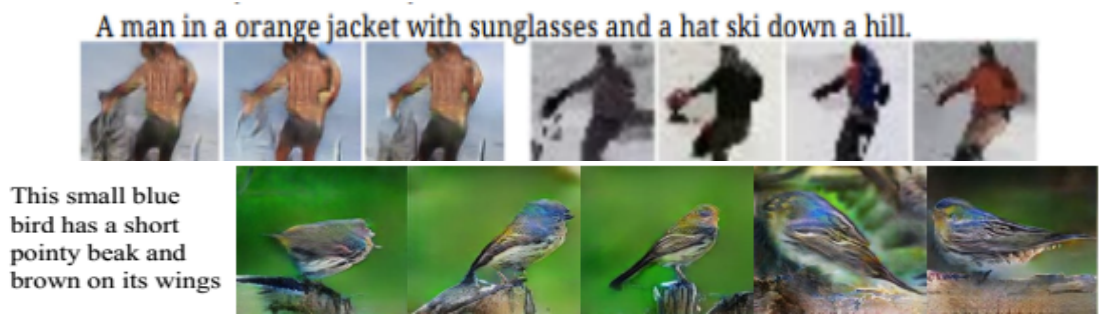

Figure 17: The upper part of the figure shows low diversity results for T2I synthesis tasks due to the mode collapse as compared to the lower part higher diversity results through FM technique. Images from [193].

## 5 Conclusion and future research directions

In this study, we have presented a survey of the GAN models, its modified classical versions, and detail analysis of various GAN applications in different domains of computer vision. Despite all these, the core idea behind this survey is to discuss the GAN model training obstacles and their potential solutions that can advance the training of GAN. The above discussion shows that GAN has the power to facilitate several new practical applications in many other domains, including those we have discussed above, like image, audio, and video in the future. Despite GAN's significant success, the architecture of GAN suffers due to unstable training. Thus, we discussed several training techniques had recommended by different researchers to stabilize training and have fixed some previous limitations for the generation of highly-realistic looking data. Irrespective of the significant progress of GAN models in recent years, several issues remaining to be dealt with for future research such as:

— **Discrete data:** GAN can generate continuous value data but cannot generate discrete value data (words, characters, bytes) directly on which natural language processing (NLP) applications heavily based; thus, GAN capability for NLP applications becomes limited. Different researchers proposed different methods such as WGAN-GP [24] modeled discrete data with a continuous generator, Boundary-Seeking GAN (B-GAN) [222] introduced a new method for the training of GAN with discrete data,

Maximum-Likelihood GAN (Mali-GAN) [223] purposed a new objective function for the discriminator network, and Adversarially Regularized AE (ARAE) [224] proposed a new method for training a discrete structure. This appealing area deserves more effort to be done.

— **Training instability:** A stable training is essential to reach Nash-equilibrium [191] for the generator to capture the real distribution of actual examples, but it is still proved difficult for both $G$ and $D$ to find a saddle point. There are early attempts towards this research direction such as WGAN [23], WGAN-GP [24], WGAN Lipschitz Penalty (WGAN-LP) [225], Feature Matching (FM) [196], Multi-Batch Discriminator (MBD) [196], One-Sided Label Smoothing [196], Historical Averaging (HA) [196], Two Time-scale Update Rule (TTUR) [198], and Spectral Normalization GAN (SN-GAN) [205] for more stable training. Training instability is another interesting future issue because more solutions should be required to make GAN training more stable and converge to Nash-equilibrium.

— **Model evaluation:** Evaluation of GAN models is one of the directions that still need to be addressed in the future because it is still difficult to evaluate which method(s) is better than other methods. Despite the early efforts towards that are broadly used at current such as Inception Scores (IS) [196], Mode Score (MS) [225], Kernel Inception Distance (KID) [226], Fréchet Inception Distance (FID) [198], Multi-Scale Structural Similarity (MS-SSIM) [227], Classifier Two-Sample Tests (C2ST) [228], Wasserstein Critic [27], and Maximum Mean Discrepancy (MMD) [229], there are no standard consensus parameters for fair model-to-model comparison. Thus, the development of better and universal quantitative evaluation metrics for this scenario still requires future research.

— **Model collapse:** As discussed in Section 4.1.3, show that GAN does suffer mode collapse, i.e., the situation in which the generator generates multiple samples with similar properties or minimal diversity amongst generated samples. Therefore, to increase the diversity of the generated samples, different solutions have also been proposed such as WGAN [23], Multi-Batch Discriminator (MBD) [196], Energy-Based GAN (EBGAN) [22]), Unrolled GAN (UGAN) [197], Deep Regret Analytic GAN (DRAGAN) [45], Adaptive GAN(AdaGAN) [231], Mode Regularized GAN (MRGAN) [232], Multi-Agent Diverse GAN (MAD-GAN) [233] that show how these methods solve mode collapse and increases the diversity of the generated samples with higher visual quality. But, the development of better training algorithm for this scenario still requires future research.

— **Others:** There are other research problems for GANs, such as disease prediction (detailed in Section 3.1.13), Nash- equilibrium (detailed in Section 4.1.1) and vanishing gradient(detailed in Section 4.1.3).

We hope this survey will help the readers to get a comprehensive overview of Generative Adversarial Networks.

Table 7: Summary of training techniques with their pros and cons to improve GAN stability

| Technique | Reference | Pros | Cons |
|---|---|---|---|
| Feature Matching | [89] | 1. Improve the training stability of GAN | 1. It has no fixed point where G exactly matches |
| Unrolled GAN | [94] | 1. Reduce mode collapse<br>2. Improve the training stability of GAN | 1. Higher computational cost<br>2. Generated images are of low quality. |
| Minibatch Discrimination | [89] | 1. Reduce mode collapse<br>2. Generate visually appealing samples very quickly. | 1. Not s strong classifier |
| Two Time-scale Update Rule | [95] | 1. Speed-up the training process<br>2. Through TTUR, models produce better results in small time | 1. Handling of separate learning parameters is difficult task |
| Self-Attention GAN | [96] | 1. Great performance on multi-class image generation | 1. Difficult implementation |
| Relativistic GAN | [62] | 1. Solve mode collapse problem<br>2. Solves vanishing gradient problem | 1. Lack of mathematical implications.<br>2. No standard rules to achieve the best performance through the inclusion of. |
| Label-smoothing | [100] | 1. Reduce the vulnerability of GAN | 1. Better for modest size training sets |
| Hybrid Model. | [49] | 1. Power of different models combined<br>2. Handling of complex situation well | 1. Complexity of the model increase<br>2. Requires more computational power |
| Sampling GAN | [101] | 1. Prevents diverging from a model's prior distribution<br>2. Produces sharper samples. | 1. Hard implementation |
| Proper Optimizer | [105] | 1. Minimize learning error rate,<br>2. Greatly improves the stability of the GANs | 1. No standard mechanism to decide which optimizer best fit for which task |
| Batch Normalization | [106] | 1. Reduces the training time<br>2. Reduce internal co-variant shift(ICS) | 1. Efficiency totally dependent on mini-batch size |
| Spectral Normalization | [113] | 1. Simple implementation<br>2. Solve mode collapse and vanishing gradient problems<br>3. Improve the training stability of GAN<br>4. The computational cost is also relatively small. | 1. Requires large image data for better performance |
| Add Noise to Inputs | [114] | 1. Stabilize GAN training | 1. Introduces significant variance in the parameter estimation process with high |
| Train with Labels | [115] | 1. Generate good quality images<br>2. More control on required result | 1. More computational power is required |
| Alternative Loss Functions(LSGAN). | [116] | 1. Simple implementation<br>2. Remove vanishing gradient problem<br>3. Improve the training stability of the GANs models | 1. Generated images are of low quality |
| Gradient Penalty. | [117] | 1. Better convergence power<br>2. Improve the training stability of the GANs models<br>3. Able to use deeper GAN model | 1. Cannot use batch normalization technique |

**ACKNOWLEDGMENTS**

I am thankful to my supervisor Prof Xi Li. I am incredibly pleased for the time he has invested in teaching everything this survey required in resolving issues that arose and for being such a good mentor.